\pdfoutput=1
\documentclass[10pt,twocolumn,letterpaper]{article}

\usepackage[pagenumbers]{cvpr}

\usepackage{graphicx}
\usepackage{amsmath}
\usepackage{amssymb}
\usepackage{booktabs}
\usepackage{multirow}
\usepackage{siunitx}
\usepackage{xspace}
\usepackage{cuted}

\definecolor{cvprblue}{rgb}{0.21,0.49,0.74}
\usepackage[pagebackref,breaklinks,colorlinks,allcolors=cvprblue,pdftitle={Mind the Band: The Hidden Frequency Variable in Geometric RoPE},pdfauthor={Seonghyun Jin, Youngmin Kim, Jong Chul Ye}]{hyperref}
\usepackage[capitalize]{cleveref}

\newcommand{\dB}{\,\text{dB}\xspace}

\title{Mind the Band: The Hidden Frequency Variable in Geometric RoPE}

\newcommand\blfootnote[1]{\begingroup\renewcommand\thefootnote{}\footnote{#1}\addtocounter{footnote}{-1}\endgroup}
\author{
Seonghyun Jin$^{*}$ \qquad Youngmin Kim$^{*}$ \qquad Jong Chul Ye$^{\dagger}$ \\[3pt]
{\normalsize Kim Jaechul Graduate School of Artificial Intelligence, KAIST} \\[2pt]
{\normalsize \url{https://bispl-wm.github.io/MindTheBand/}}
}

\begin{document}
\maketitle
\blfootnote{$^{*}$Equal contribution.\quad $^{\dagger}$Corresponding author.}
\begin{strip}
  \centering
  \vspace{-3.6em}
  \includegraphics[width=\linewidth]{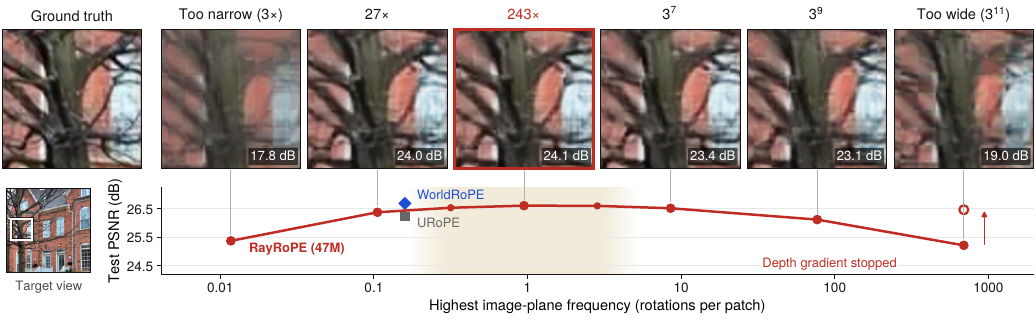}
  \captionsetup{hypcap=false}
  \captionof{figure}{\textbf{The frequency band governs geometric RoPE.}
  Varying only single-ray RayRoPE's frequency range (47M) shifts novel-view renderings from blurred ($3\times$) to sharp ($243\times$) before degrading again at $3^{11}$---ranges that existing defaults trigger implicitly as head size changes.
  Blocking high-frequency depth gradients at $3^{11}$ restores accuracy (open circle; shading: $0.3$--$3$ rotations/patch, $10$--$90$/image).}
  \captionsetup{hypcap=true}
  \label{fig:teaser}
  \vspace{-0.5em}
\end{strip}

\begin{abstract}
Geometric rotary position encodings (RoPE) incorporate 3D camera geometry into multi-view attention by rotating token features according to viewing rays, depths, and reprojected image coordinates.
Each coordinate is rotated across a progression of frequencies---its \emph{frequency band}---yet existing implementations leave this band to inherited defaults, either retaining standard 2D RoPE's top frequency or letting the frequency span grow exponentially with the attention head dimension.
We show that this unstandardized setting can reverse benchmark rankings.
In novel view synthesis on RealEstate10K, RayRoPE matches the strongest baseline in the 47M model (144-channel heads), yet trails by $1.2\dB$ in the 25M model (48-channel heads)---and exhibits the same reversal when varying head size alone at fixed model width.
Populating a $243\times$ range with four frequencies instead of the $3\times$ default at 48 channels recovers $0.9\dB$ of that gap.
Across view synthesis and two-view depth estimation, overly narrow bands consistently blur correspondence, whereas all three projection-based encodings perform reliably when their top frequency lies within $0.3$--$3$ rotations per 8-pixel patch ($10$--$90$ rotations across $256\times256$ RealEstate10K).
At the opposite extreme, overly wide bands degrade RayRoPE through a forward--backward asymmetry: averaging rotations over a predicted depth interval damps high frequencies in the forward pass, yet leaves their gradients with respect to predicted depth undamped.
Blocking only those high-frequency depth gradients recovers 90--93\% of the lost PSNR, showing that forward smoothing does not guarantee backward stability.
Guided by this analysis, we distill practical guidelines---setting bands in image coordinates, populating spans in small heads, and avoiding unconditional same-view damping---and advocate reporting and matching frequency bands in future multi-view benchmarks.
\end{abstract}

\begin{figure*}[t]
  \centering
  \includegraphics[width=\textwidth]{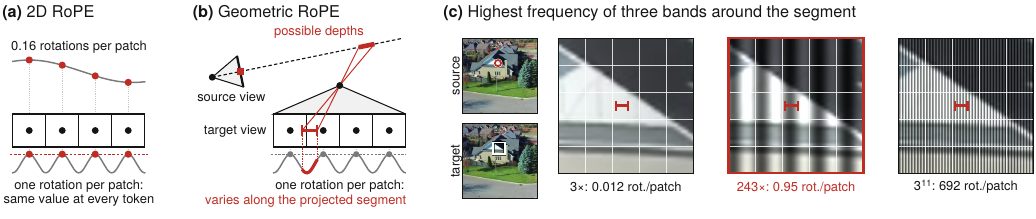}
  \caption{\textbf{From discrete patch indices to continuous projected segments.}
  (a)~In 2D RoPE, tokens sit at integer patch centers, where a frequency of one rotation per patch gives every token the same phase.
  (b)~Across cameras, geometric RoPE projects a token's 3D viewing ray into the other view: the reprojected position is continuous, and unknown depth spreads it along a segment.
  (c)~Target crop around the projected segment ($0.45$ patches, red) for a $\pm15\%$ depth interval of the marked source pixel, overlaid with the top frequency of three bands: $3\times$ barely varies across patches, $243\times$ separates sub-patch offsets, and $3^{11}$ turns ${\sim}300$ times inside the segment (drawn $77\times$ coarser).}
  \label{fig:concept}
\end{figure*}

\section{Introduction}
\label{sec:intro}

Vision transformers~\cite{vaswani2017attention,dosovitskiy2021vit} commonly encode 2D token positions with Rotary Position Embedding (RoPE)~\cite{su2024roformer}, rotating query and key channels across a geometric progression of frequencies from slow image-wide waves to fast local ones.
Within a single image, tokens sit on an integer grid of patch centers, so frequencies above half a rotation per patch alias (\cref{fig:concept}(a)), and standard 2D RoPE schedules start at $0.16$ rotations ($1$ radian) per patch below that limit~\cite{heo2024rope}.
Multi-view tasks such as novel view synthesis~\cite{sajjadi2022srt,jin2025lvsm} and 3D reconstruction~\cite{zhang2024gslrm,wang2024dust3r,wang2025vggt}, however, must relate tokens across different cameras, where matching pixels depend on 3D scene depth as well as camera pose.
Going beyond relative camera transforms~\cite{miyato2024gta,li2025prope}, projection-based geometric RoPE methods project each token's 3D viewing ray into the attending camera so tokens on the same 3D surface receive matching coordinates in attention (\cref{fig:concept}(b)).
Unknown depth along each ray is represented via fixed depth planes (URoPE~\cite{xie2026urope}), a predicted depth interval (RayRoPE~\cite{wu2026rayrope}), or predicted 3D Gaussians (WorldRoPE~\cite{worldrope2026}).

Projecting continuous 3D rays rather than indexing a 2D patch grid changes what a coordinate's set of rotation frequencies---its \emph{frequency band}---must capture (\cref{fig:concept}(c)).
Because reprojected points can land anywhere inside a patch, frequencies near one rotation per patch no longer repeat at every token and instead separate sub-patch offsets across views, while depth uncertainty spreads each projection along a segment over which much faster frequencies cancel out.
Yet existing implementations do not standardize the band around this geometry.
URoPE and WorldRoPE carry over 2D RoPE's $0.16$-rotations/patch ceiling unchanged, whereas RayRoPE, CRePE~\cite{jin2026crepe}, and G-ray's RayRoPE variant~\cite{gray2026} fix the lowest frequency and multiply by three per channel pair, causing the span to grow exponentially with the head dimension $d$---from $3\times$ at $d=48$ to $243\times$ at $d=144$ (\cref{fig:teaser}).

This overlooked convention can overturn empirical rankings.
Holding model width fixed on RealEstate10K while varying only head size (\cref{fig:sweep}(a)), RayRoPE trails URoPE by $0.7\dB$ and WorldRoPE by $1.2\dB$ at $d=48$, yet overtakes URoPE and approaches WorldRoPE as heads widen; holding RayRoPE's range at $243\times$ removes this head-size dependence almost entirely.
Benchmark rankings can thus hinge on an unreported frequency setting unrelated to the geometry being compared, echoing hidden confounders across machine learning~\cite{lucic2018gans,musgrave2020reality,henderson2018matters,melis2018evaluation}.

In this paper, we systematically study how frequency bands govern geometric RoPE across novel view synthesis and two-view depth estimation.
Changing the band alone shifts RayRoPE by $1$ to $2\dB$; across seven structural choices separating published encodings, only depth-based reprojection has a comparable impact (\cref{fig:effects}).
On $256\times256$ images with 8-pixel patches, RayRoPE stays within $0.1\dB$ of its peak when its top frequency lies between $0.3$ and $3$ rotations per patch (\cref{fig:teaser}); URoPE and WorldRoPE are equally stable there, whereas compressing any of the three to $3\times$ costs $0.6$ to $1.2\dB$.

We then analyze why bands fail at both extremes (\cref{fig:concept}(c), \cref{fig:mechanism}(a)).
When the band is too narrow, neighboring patches receive nearly identical rotations, starving attention of local positional contrast and blurring rendered views.
When it is far too wide, failure stems from the backward pass: averaging rotations over RayRoPE's predicted depth interval damps fast frequencies forward, yet differentiating that average with respect to the interval's center cancels this damping at the sharp endpoints.
Frequencies that barely affect outputs thus still inject large, rapidly oscillating gradients into predicted depth, and blocking only those gradients recovers 90--93\% of the lost PSNR.

Finally, we distill four practical guidelines (\cref{sec:discussion}): set bands directly in image units rather than via $d$, keep and populate the span in small heads, avoid unconditional smoothing within the same view where image coordinates do not depend on depth, and report and match bands in multi-view benchmarks.

\section{Related Work}
\label{sec:related}

\paragraph{Camera encodings for multi-view transformers.}
Early multi-view models injected camera geometry at the input via pixel-aligned ray maps~\cite{sajjadi2022srt,gao2024cat3d}, such as world-frame Pl\"ucker rays in LVSM~\cite{jin2025lvsm}.
Recent methods move geometry into attention along two main lines.
Pose- and frustum-transform encodings transform queries and keys by relative camera geometry without depth reprojection (GTA~\cite{miyato2024gta}, CaPE~\cite{kong2024eschernet}, PRoPE~\cite{li2025prope}, and extensions to general cameras, scaling, and video~\cite{ucpe2025,hetero2026,dppe2026}; cf.\ additive Pl\"ucker-ray terms in SCoPE~\cite{scope2025}, epipolar disparity anchors in MeRoPE~\cite{merope2026}, and local homographies in HRoPE~\cite{hrope2026}).
Projection-based encodings instead reproject 3D points along each token's viewing ray at fixed depths (URoPE~\cite{xie2026urope}), over a predicted depth interval (RayRoPE~\cite{wu2026rayrope}, also hosting G-ray~\cite{gray2026} and extended in CRePE~\cite{jin2026crepe}), or over a predicted world-space Gaussian mixture (WorldRoPE~\cite{worldrope2026}; cf.\ POD-RoPE~\cite{podrope2026}).
Rather than proposing another parameterization, we study a shared axis left unstandardized across comparisons: each coordinate's band.

\paragraph{Frequencies of positional encodings.}
In language models, interpolating positions or rescaling frequencies extends context length~\cite{chen2023pi,peng2024yarn,liu2024scaling,men2024base}, while positional heads rely on the highest frequencies for sharp local alignment~\cite{barbero2025round}.
Vision and multimodal transformers split head channels across spatial and temporal axes~\cite{heo2024rope,hope2025,multimodalpe2025}, learn mixed rotations~\cite{ostmeier2025liere,harope2025}, or scale frequencies with resolution~\cite{lu2024fit}.
In neural fields, Fourier bandwidth governs representable detail~\cite{mildenhall2020nerf,tancik2020fourier,rahaman2019spectral}: Mip-NeRF~\cite{barron2021mipnerf} and Zip-NeRF~\cite{barron2023zipnerf} prefilter over pixel cones to anti-alias fine scales, while BARF~\cite{lin2021barf}, noting that point-code gradients grow as $\omega$, and related works~\cite{park2021nerfies,yang2023freenerf} anneal or mask high frequencies.
Within geometric RoPE, prior works incorporate uncertainty smoothing~\cite{wu2026rayrope,worldrope2026} and multi-frequency or frequency-scale ablations~\cite{wu2026rayrope,gray2026}, yet leave the band unstandardized across backbones and baselines---and overlook how interval averaging damps high frequencies as $1/\omega$ forward while sharp endpoints cancel that damping in backward depth gradients.

\section{Frequency Bands in Geometric RoPE}
\label{sec:bands}

\subsection{From 2D Patch Grids to 3D Ray Projections}

Standard RoPE~\cite{su2024roformer} encodes a scalar position $x$ by grouping feature channels into complex pairs and multiplying the $k$-th pair by $e^{i\omega_k x}$, so the inner product between a query at $x$ and a key at $x'$ depends only on their offset $x - x'$.
The frequencies $\omega_0 < \dots < \omega_{n-1}$ form a geometric progression: low frequencies distinguish distant tokens without wrapping around $2\pi$, while high frequencies separate nearby tokens.

In a 2D vision transformer, $x$ is an integer patch index (\cref{fig:concept}(a)).
Because shifting $\omega$ by a whole turn per patch leaves $e^{i\omega x}$ identical at every integer center, frequencies above $0.5$ rotations per patch alias to lower ones; conventional fixed 2D axial RoPE schedules~\cite{heo2024rope} (as in CroCo v2's 2D RoPE~\cite{weinzaepfel2023crocov2}) start at $1$ radian ($1/(2\pi) \approx 0.16$ rotations) per patch, below that limit.

Geometric RoPE methods instead evaluate rotations at continuous camera-derived coordinates, often rotating values and outputs alongside queries and keys.
\Cref{fig:concept}(b) illustrates how projection-based methods relate two cameras: a token in a source view knows the 3D viewing ray through its patch, but not the surface depth along that ray.
URoPE~\cite{xie2026urope} places fixed depth planes along the ray (assigning different planes across head groups) and projects them into the attending view.
RayRoPE~\cite{wu2026rayrope} predicts a per-token depth and uncertainty in each layer, transforms the camera center and resulting ray segment into the attending camera, and averages rotations uniformly over the projected interval $[\mu-\Delta, \mu+\Delta]$.
WorldRoPE~\cite{worldrope2026} predicts a mixture of 3D Gaussians along the ray---incorporating each source patch's spatial footprint---and takes the closed-form expected rotation after splatting them into the attending view.
Across views, their projected coordinates are not restricted to the discrete token grid.
Within the \emph{same} view, however, every point along a ray projects back to the token's own pixel, so projected 2D image coordinates do not depend on depth there (only a separate scalar depth channel, when used, varies with depth).

\begin{figure*}[t]
  \centering
  \includegraphics[width=\textwidth]{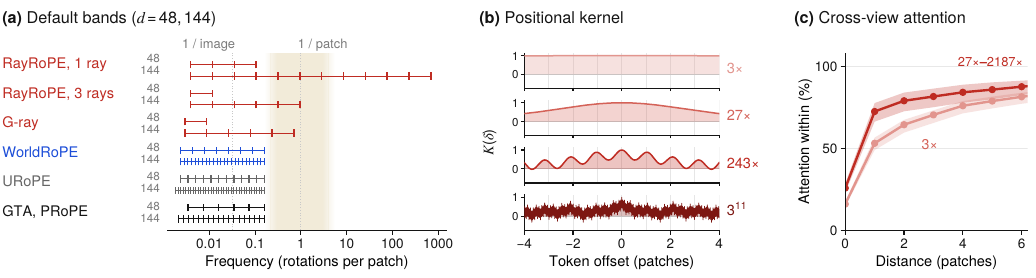}
  \caption{\textbf{Default frequency bands and spatial selectivity.}
  (a)~Image-plane bands assigned by each released rule at $d=48$ and $144$ (G-ray shown with its RayRoPE host; shading: RayRoPE's plateau).
  (b)~Unweighted positional kernel $K(\delta)$ (\cref{eq:kernel}) across patch offsets for four frequency ranges.
  (c)~Fraction of cross-view attention within a given distance of verified correspondences in 25M RayRoPE models (mean, min--max over four $3\times$ and six wider-band models).}
  \label{fig:bandchart}
\end{figure*}

If a head of dimension $d$ rotates $C$ scalar coordinates equally, each receives $n = d/(2C)$ complex pairs---its \emph{frequency band} $\{\omega_0, \dots, \omega_{n-1}\}$.
With $\omega_0$ anchored at a coarse image-wide scale, a band is governed mainly by its top frequency $\omega_{n-1}$ (or \emph{range} $\omega_{n-1}/\omega_0$ at fixed $\omega_0$).
To compare image coordinates expressed in patch indices, normalized coordinates, or ray angles, we report image-plane frequencies in \emph{rotations per patch} ($2\pi$ phase shift per patch width, or $1/32$ of a $256\times256$ image with $8\times8$ patches).
Because cross-view projections land at continuous positions, a frequency near $1$ rotation per patch separates sub-patch offsets (\cref{fig:concept}(c)), lower frequencies resolve multi-patch ambiguity (\cref{fig:bandchart}(b)), and the depth segment bounds how high a frequency helps (\cref{sec:analysis}).

\subsection{How Existing Implementations Set Bands}

\Cref{fig:bandchart}(a) compares default image-plane bands at $d=48$ and $d=144$.
URoPE and WorldRoPE implementations follow the top-down schedule of 2D axial RoPE~\cite{heo2024rope}: they fix $\omega_{n-1}$ at $1$ radian ($0.16$ rotations) per patch and fill in lower or denser frequencies as $d$ grows.

RayRoPE, CRePE, and RayRoPE-hosted G-ray adopt the opposite bottom-up rule (other G-ray hosts use top-down schedules~\cite{gray2026}): they fix $\omega_0$ ($1/8$ rotation per image in RayRoPE) and triple each step,
\begin{equation}
  \omega_k = \omega_0\, 3^{k}, \qquad k = 0, \dots, n-1,
  \label{eq:growing}
\end{equation}
so the range $3^{\,n-1}$ grows exponentially with $n = d/(2C)$ and thus with head size $d$.
RayRoPE rotates the 3D camera center plus a 2D projected direction and scalar depth per ray: the paper formulates one ray per patch ($C=6$), whereas the released code samples three ($C=12$).
Under \cref{eq:growing}, three-ray RayRoPE assigns $n=6$ frequencies spanning $243\times$ ($\omega_{n-1} = 0.95$ rotations per patch) at $d=144$, but shrinking heads to $d=48$ leaves just $n=2$ frequencies spanning $3\times$ ($0.012$ rotations per patch).
Conversely, using one ray at $d=144$ yields $n=12$ frequencies spanning $3^{11}$ ($692$ rotations per patch).

The host backbone thus silently dictates spatial scale, yet cross-method benchmarks rarely match or report the resulting bands.
Across ten published evaluations of RayRoPE-family encodings outside the original paper (supplement), only three use the $d=144$ head size for which $3^k$ was designed, four use $d=48$ (where the unchanged three-ray rule yields $3\times$), two use head sizes not divisible by that rule ($d=64$~\cite{worldrope2026}, $d=128$~\cite{merope2026}), and only two (G-ray, CRePE) report the frequencies.
Under the released three-ray rule, these backbones span $3\times$ to $2187\times$ ($d=192$), while $3^{11}$ arises in the single-ray formulation at $d=144$ and serves below to expose how overly wide bands fail (\cref{sec:analysis}).

\section{Experiments}
\label{sec:experiments}

\begin{figure*}[t]
  \centering
  \includegraphics[width=\textwidth]{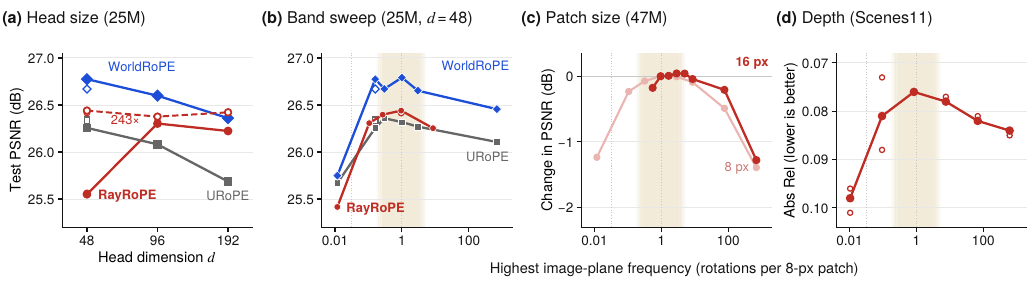}
  \caption{\textbf{Performance across head sizes and maximum frequencies} (open markers: extra seeds; shading: RayRoPE's plateau).
  (a)~Varying head count in the 25M model on RealEstate10K; holding RayRoPE at $243\times$ (dashed) removes its head-size dependence.
  (b)~Sweeping the top image-plane frequency at fixed $\omega_0$.
  (c,\,d)~Single-ray RayRoPE with 16- vs.\ 8-px patches (47M)~(c) and on two-view depth~(d).}
  \label{fig:sweep}
\end{figure*}

\subsection{Setup}

Our RealEstate10K~\cite{zhou2018re10k} view-synthesis experiments ($256\times256$ images, $8\times8$ patches) use the public WorldRoPE training framework~\cite{worldrope2026} across two standard LVSM backbones: the 25M model (width 768, 16 heads of $d=48$) and the 47M model (width 1152, 8 heads of $d=144$).
To test generalization across camera models, datasets, and tasks, we also evaluate fisheye view synthesis on ScanNet++~\cite{yeshwanth2023scannetpp}, posed two-view depth estimation with UniMatch~\cite{xu2023unimatch} on Scenes11~\cite{ummenhofer2017demon}, and a 16-run fractional-factorial ablation of seven encoding design choices (\cref{sec:other}), using a common 47M LVSM implementation for the fisheye and multi-encoding runs.
Within every comparison, models share weight initialization and batch order; repeated runs vary by at most $0.12\dB$ across stable settings (and up to $1.85\dB$ only in the unstable $3^{11}$ regime; full protocols and 95\% bootstrap intervals in the supplement).

\subsection{Head-Size Sensitivity Is Band Sensitivity}

\begin{table}[t]
  \centering
  \caption{\textbf{Novel view synthesis on RealEstate10K.}
  ``Default'': released three-ray RayRoPE at each $d$; SSIM~\cite{wang2004ssim}, LPIPS~\cite{zhang2018lpips}.
  $\Delta$: paired PSNR gap to WorldRoPE ($\pm0.03$\dB 95\% bootstrap; supplement); ``$+$ stop-gradient'' blocks high-frequency depth gradients (\cref{sec:verification}).
  Additional runs of the last three 47M rows score $26.66\dB$ ($243\times$), $23.56$--$25.41\dB$ ($3^{11}$), and $26.52\dB$ ($+$ stop-gradient; supplement).}
  \label{tab:official}
  \footnotesize
  \setlength{\tabcolsep}{1.9pt}
  \begin{tabular}{@{}lccrcccr@{}}
    \toprule
    Method & Rays & Freq. & Range & PSNR$\uparrow$ & SSIM$\uparrow$ & LPIPS$\downarrow$ & $\Delta$ \\
    \midrule
    \multicolumn{8}{@{}l}{\emph{47M model, 144-dimensional heads}} \\
    WorldRoPE & -- & 24 & $68\times$ & 26.69 & 0.844 & 0.073 & \\
    URoPE & -- & 36 & $88\times$ & 26.24 & 0.834 & 0.078 & $-$0.45 \\
    RayRoPE (default) & 3 & 6 & $243\times$ & 26.66 & 0.845 & 0.074 & $-$0.02 \\
    RayRoPE & 1 & 12 & $243\times$ & 26.61 & 0.844 & 0.075 & $-$0.08 \\
    RayRoPE & 1 & 12 & $3^{11}$ & 24.54 & 0.779 & 0.181 & $-$2.15 \\
    \quad $+$ stop-gradient & 1 & 12 & $3^{11}$ & 26.47 & 0.841 & 0.077 & $-$0.22 \\
    \midrule
    \multicolumn{8}{@{}l}{\emph{25M model, 48-dimensional heads}} \\
    WorldRoPE & -- & 8 & $68\times$ & 26.77 & 0.847 & 0.072 & \\
    URoPE & -- & 12 & $68\times$ & 26.26 & 0.834 & 0.079 & $-$0.52 \\
    RayRoPE (default) & 3 & 2 & $3\times$ & 25.55 & 0.817 & 0.089 & $-$1.22 \\
    RayRoPE & 1 & 4 & $3\times$ & 25.42 & 0.813 & 0.092 & $-$1.36 \\
    RayRoPE & 1 & 4 & $27\times$ & 26.31 & 0.835 & 0.080 & $-$0.47 \\
    RayRoPE & 1 & 4 & $68\times$ & 26.40 & 0.838 & 0.079 & $-$0.38 \\
    RayRoPE & 1 & 4 & $243\times$ & 26.44 & 0.840 & 0.079 & $-$0.33 \\
    RayRoPE & 1 & 4 & $2187\times$ & 26.25 & 0.835 & 0.080 & $-$0.52 \\
    \bottomrule
  \end{tabular}
\end{table}

Under each method's default rule (\cref{tab:official}), RayRoPE matches WorldRoPE and outperforms URoPE in the 47M backbone ($d=144$), yet drops by more than $1\dB$ in the 25M backbone ($d=48$) and trails both.
This is the RealEstate10K ranking reported by the WorldRoPE and URoPE papers~\cite{worldrope2026,xie2026urope}, the former evaluated at $d=48$ and the latter released with a $d=48$ configuration.

To separate head dimension from model capacity, \cref{fig:sweep}(a) varies only the head count ($16, 8,$ and $4$ heads, giving $d=48, 96,$ and $192$) at fixed 25M width.
Whereas WorldRoPE and URoPE decline gently as heads become fewer and wider, default RayRoPE jumps sharply from $d=48$ to $96$ (overtaking URoPE) and approaches WorldRoPE at $d=192$.
When we instead hold RayRoPE's frequency range fixed at $243\times$ across all three head sizes (using one ray per patch at $d=48$; dashed curve in \cref{fig:sweep}(a)), its PSNR stays flat within $0.06\dB$, showing that its apparent head-size sensitivity is in fact band sensitivity.

\subsection{Mapping the Frequency Band}

To vary the band independently of $d$, we keep $\omega_0$ fixed and scale the geometric step ratio so $\omega_{n-1}/\omega_0$ spans a chosen range.
At $d=48$, three-ray RayRoPE leaves only $n=2$ frequencies per coordinate; stretching two lone frequencies across $243\times$ gains only $0.2\dB$ whether using one ray or three (supplement), so a wide range also needs intermediate frequencies to populate it.
Using one ray per patch ($C=6$) provides $n=4$ frequencies per coordinate (\cref{tab:official}, bottom).
Kept at $3\times$, single-ray RayRoPE performs no better than the three-ray default, ruling out ray count itself as the bottleneck.
Widening those four frequencies to $243\times$ ($\omega_{n-1}\approx 1$ rotation per patch) recovers $0.9\dB$ of RayRoPE's $1.2\dB$ deficit---surpassing URoPE and narrowing the gap to WorldRoPE to $0.3$--$0.4\dB$ across $68\times$ and $243\times$.
Conversely, at 47M ($d=144$, $n=12$; \cref{tab:official}, top), single-ray RayRoPE under the default $3^k$ rule spans $3^{11}$ and drops to $23.56$--$25.41\dB$ across five runs, whereas rescaling those twelve frequencies to $243\times$ ($26.61$ and $26.66\dB$) matches the three-ray default within $0.05\dB$.

\begin{figure*}[t]
  \centering
  \includegraphics[width=\textwidth]{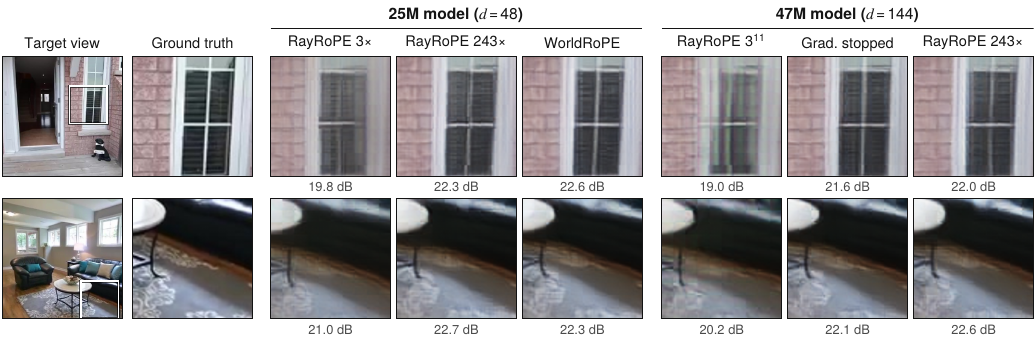}
  \caption{\textbf{Qualitative comparison on RealEstate10K} (full-view PSNR below each crop).
  Narrowing RayRoPE's band to $3\times$ blurs fine detail, whereas widening it to $3^{11}$ introduces colored streaks and blurred texture that disappear when high-frequency depth gradients are stopped.}
  \label{fig:qual}
\end{figure*}

Plotted against the top image-plane frequency (\cref{fig:teaser,fig:sweep}(b)), 25M and 47M RayRoPE trace nearly identical curves, staying within $0.1\dB$ of their peak across a broad plateau from $0.3$ to $3$ rotations per 8-pixel patch ($10$--$90$ rotations across the image, with $243\times$ at $0.95$ rotations per patch near its center).
Doubling patches to $16\times16$ pixels (\cref{fig:sweep}(c); supplement) is closer to holding rotations per image fixed ($1.9$--$17$ per 16-px patch) than rotations per patch, though the broad plateau leaves exact scaling open.
Below this plateau, accuracy falls steeply and renderings blur; far above it, accuracy drops and renderings show streaks and blur (\cref{fig:qual}).
Varying one coordinate group at a time (supplement) shows that the 2D projected ray direction drives both regimes, accounting for 93\% of narrow-band loss and 84\% of wide-band loss, whereas widening camera center or scalar depth alone has minimal effect.

Sweeping the image-plane bands of URoPE and WorldRoPE (\cref{fig:sweep}(b); supplement) separates shared behavior from RayRoPE-specific behavior.
Compressing either encoding to $3\times$ degrades PSNR by $0.6$--$1.2\dB$ across 25M and 47M, so narrow bands hurt all three encodings.
Raising $\omega_{n-1}$ from their $0.16$-rotation default to $0.3$, $1$, or $3$ rotations/patch leaves both within $0.02\dB$ of their default seed range at 25M (supplement): 2D RoPE's $1$-radian/patch ceiling already lies near the effective regime on $256\times256$ images, though continuous reprojection does not require it.
At the extreme $3^{11}$ band, however, URoPE and WorldRoPE drop by at most $0.2$--$0.3\dB$ at 25M and under $0.1\dB$ at 47M, isolating severe wide-band failure to RayRoPE.

\subsection{Other Datasets, Tasks, and Design Choices}
\label{sec:other}

\begin{table}[t]
  \centering
  \caption{\textbf{Novel view synthesis on ScanNet++} (fisheye test PSNR) across head dimensions $d=144, 48,$ and $96$ (width 1152 at $d=144$, 768 otherwise), corresponding to ranges of $243\times$, $3\times$, and $27\times$ under \cref{eq:growing} for the first three rows (G-ray on its RayRoPE host).
  $^\ast$Image-plane coordinates only; baseline details in supplement.}
  \label{tab:scannet}
  \footnotesize
  \setlength{\tabcolsep}{5.2pt}
  \begin{tabular}{@{}lccc@{}}
    \toprule
    Method & $d=144$ & $d=48$ & $d=96$ \\
    \midrule
    RayRoPE~\cite{wu2026rayrope} & 23.76 & 22.40 & 23.37 \\
    G-ray (RayRoPE)~\cite{gray2026} & 23.81 & 22.48 & 23.27 \\
    CRePE~\cite{jin2026crepe} & \textbf{23.84} & 22.45 & 23.23 \\
    WorldRoPE$^\ast$~\cite{worldrope2026} & 23.25 & \textbf{23.47} & 23.23 \\
    URoPE~\cite{xie2026urope} & 23.16 & 23.27 & 23.04 \\
    Pl\"ucker~\cite{jin2025lvsm} + 2D RoPE & 21.39 & 21.59 & 21.32 \\
    \bottomrule
  \end{tabular}
\end{table}

\paragraph{Fisheye ScanNet++ and two-view depth.}
ScanNet++ exhibits the same ranking reversal (\cref{tab:scannet}): moving from $d=144$ to $48$ drops RayRoPE, RayRoPE-hosted G-ray, and CRePE by over $1.3\dB$---from first place to behind WorldRoPE$^\ast$ and URoPE.
At the width of the $d=48$ model, heads of $d=96$ already recover $0.8$ to $1.0\dB$, and rescaling the band to $243\times$ restores the lead (supplement).
Likewise, in posed two-view depth estimation with UniMatch on Scenes11 (\cref{fig:sweep}(d); supplement), mean depth error (Abs Rel) across two seeds is lowest at $243\times$; both narrow ($3\times$) and overly wide ($3^{11}$) bands increase error in every seed, with a steeper penalty on the narrow side just as in view synthesis.

\begin{figure}[t]
  \centering
  \includegraphics[width=\columnwidth]{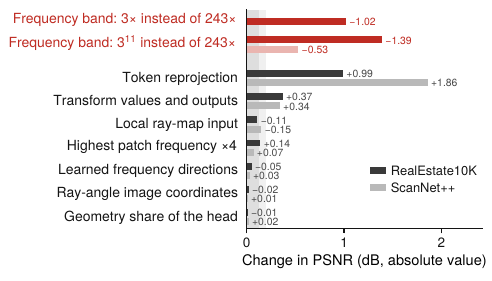}
  \caption{\textbf{Effect of the frequency band versus seven design choices} (absolute change in PSNR; shading: single-seed noise).
  Moving RayRoPE from $243\times$ to its $3\times$ or $3^{11}$ default (red; supplement) matters more than six of the seven choices; only depth-based reprojection itself is comparable.}
  \label{fig:effects}
\end{figure}

\paragraph{Band versus seven structural choices.}
To put band effects in perspective, \cref{fig:effects} compares them with a 16-run fractional-factorial study of seven design choices separating published encodings (listed in the figure; supplement).
Four of the seven fall within single-seed noise, rotating values/outputs adds $\sim 0.3\dB$, and only depth-based reprojection ($+1.0$ to $+1.9\dB$) rivals choosing a proper band.

\section{Why Bands Fail at Both Extremes}
\label{sec:analysis}

\subsection{Narrow Bands Blur Spatial Selectivity}

For identical unit-norm query and key channels, rotating pairs separated by coordinate offset $\delta$ modulates their attention score via the content-independent diagnostic kernel~\cite{su2024roformer,men2024base}
\begin{equation}
  K(\delta) = \frac1n \sum_{k=0}^{n-1} \cos(\omega_k\,\delta),
  \label{eq:kernel}
\end{equation}
shown in \cref{fig:bandchart}(b).
Under a $3\times$ band ($\omega_{n-1}=0.012$ rotations per patch), even the fastest frequency completes less than half a turn across the entire image, leaving $K(\delta)$ nearly flat across neighboring patches compared with the sharp peak at $243\times$.
Positional rotations alone then provide little local discrimination across adjacent patches, leaving fine alignment largely to token features and blurring rendered views (\cref{fig:qual}).
Trained cross-view attention on verified SIFT correspondences agrees (\cref{fig:bandchart}(c)): attention peaks near the true match in every model, yet only 53\% of attention weight concentrates within one patch under $3\times$, versus 72\% under wider bands.

\subsection{Wide Bands Leave Depth Gradients Undamped}

At first glance, extreme ranges such as $3^{11}$ should be harmless, as RayRoPE averages rotations over predicted depth uncertainty~\cite{wu2026rayrope}: on projected interval $[\mu-\Delta, \mu+\Delta]$, uniform averaging yields
\begin{equation}
  \Phi(\mu) = \frac{1}{2\Delta}\!\int_{\mu-\Delta}^{\mu+\Delta} e^{i\omega x}\,dx
  = \underbrace{e^{i\omega\mu}}_{\text{center phase}}\,
    \underbrace{\frac{\sin(\omega\Delta)}{\omega\Delta}}_{\text{damps as } 1/\omega}.
  \label{eq:box}
\end{equation}
When $\omega\Delta \gg 1$, the forward amplitude $|\Phi(\mu)| \le 1/(\omega\Delta)$ shrinks toward zero, so ultra-high frequencies barely affect attention outputs and might appear harmless.

\begin{figure*}[t]
  \centering
  \includegraphics[width=\textwidth]{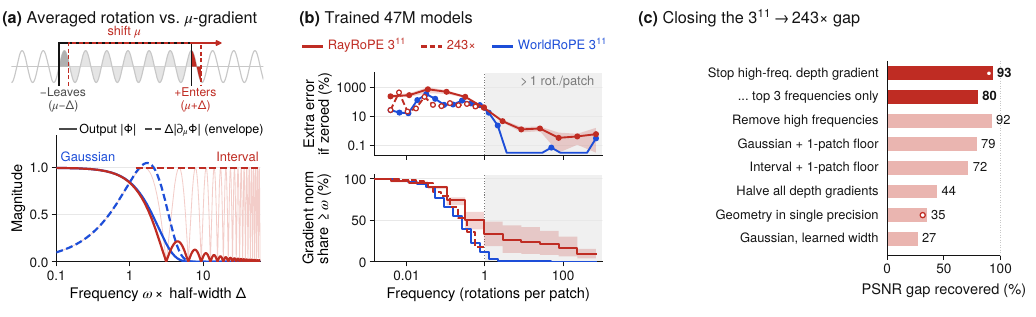}
  \caption{\textbf{Why overly wide bands harm RayRoPE} (47M models; high frequencies: $>1$ rotation/patch).
  (a)~Uniform averaging over $[\mu-\Delta, \mu+\Delta]$ damps the forward output $|\Phi|$ as $1/\omega$, but shifting $\mu$ leaves boundary changes undamped (\cref{eq:box-grad}; dashed: its envelope), unlike a smooth Gaussian window.
  (b)~Zeroing single high frequencies on keys at test time barely changes rendering error (\emph{top}), yet they carry on average one third of RayRoPE's summed per-frequency depth-gradient norms (\emph{bottom}; $3^{11}$: mean and min--max across three runs).
  (c)~Share of the $3^{11}\!\to\!243\times$ PSNR gap recovered by each intervention on 47M RayRoPE (white dot: second seed).}
  \label{fig:mechanism}
\end{figure*}

Backpropagating to the predicted interval center $\mu$, however, tells the opposite story.
Differentiating \cref{eq:box} with respect to $\mu$ pulls an $i\omega$ factor out of $e^{i\omega\mu}$ that \emph{cancels} the $1/\omega$ forward damping:
\begin{equation}
  \begin{aligned}
    \frac{\partial \Phi}{\partial \mu}
    &= \underbrace{i\omega\,e^{i\omega\mu}}_{\omega\text{ pops out}}\cdot
       \frac{\sin(\omega\Delta)}{\omega\Delta}
     = i\,e^{i\omega\mu}\,\frac{\sin(\omega\Delta)}{\Delta}, \\[3pt]
    \Bigl|\frac{\partial \Phi}{\partial \mu}\Bigr|
    &= \underbrace{\frac{|\sin(\omega\Delta)|}{\Delta}}_{\text{no } 1/\omega \text{ decay}}.
  \end{aligned}
  \label{eq:box-grad}
\end{equation}
Unlike the forward output, $|\partial_\mu \Phi|$ \emph{does not decay} as $\omega \to \infty$ (\cref{fig:mechanism}(a), red).
Geometrically, the cycles inside $[\mu-\Delta, \mu+\Delta]$ cancel, so shifting $\mu$ changes the integral only at its two sharp endpoints, where one value enters and one leaves (\cref{fig:mechanism}(a), top): $\partial_\mu\Phi = \bigl(e^{i\omega(\mu+\Delta)} - e^{i\omega(\mu-\Delta)}\bigr)/(2\Delta)$, however fast $e^{i\omega x}$ oscillates in between.
Thus even when a high frequency vanishes in the forward average, its $\mu$-derivative retains a $1/\Delta$ envelope whose phase rotates every half-wavelength as $\mu$ shifts (and $\partial_\Delta \Phi$ likewise fails to decay; supplement).
In attention, each attention pair multiplies the two tokens' coefficients, so this $O(1)$ derivative survives whenever the paired token is unattenuated---e.g., in its own camera or with a narrow interval---and scales as $O(1/\omega)$ when both are wide; since full gradients also pass through $\Delta$ and projection (supplement), \cref{sec:verification} tests by intervention whether they drive the collapse.

Smooth windows avoid this cancellation.
For a Gaussian window of width $\sigma$, as in WorldRoPE~\cite{worldrope2026}, super-exponential forward damping overwhelms the linear $\omega$ factor from differentiation (\cref{fig:mechanism}(a), blue):
\begin{equation}
  \Phi_{\text{G}}(\mu) = e^{i\omega\mu}\,e^{-\omega^2\sigma^2/2}
  \;\Longrightarrow\;
  \Bigl|\frac{\partial \Phi_{\text{G}}}{\partial \mu}\Bigr|
  = \underbrace{\omega\,e^{-\omega^2\sigma^2/2}}_{\to\,0\text{ for }\omega \gg 1/\sigma}.
  \label{eq:gauss-grad}
\end{equation}
If $\sigma$ itself is predicted, confident tokens with small $\sigma$ can still admit high frequencies; convolving the window with a fixed query-plane Gaussian floor of variance $\sigma_0^2$ bounds $|\partial_\mu \Phi| \le \omega e^{-\omega^2\sigma_0^2/2}$ regardless of predicted width (supplement), though any fixed floor within the same view also damps depth-free channels.

\subsection{Empirical and Causal Verification}
\label{sec:verification}

\paragraph{Forward attenuation vs.\ gradient load.}
Probing trained 47M models (\cref{fig:mechanism}(b)) matches \cref{eq:box,eq:box-grad}.
In $3^{11}$ RayRoPE, zeroing any single key frequency above one rotation per patch at test time barely alters rendering error, yet those six frequencies carry 19\%--49\% (mean one third) of the summed per-frequency depth-gradient norms across three training runs.
By contrast, in $243\times$ RayRoPE the frequencies that drive outputs also dominate the gradient norms, and in $3^{11}$ WorldRoPE frequencies above three rotations per patch contribute under 1\%.

\paragraph{Causal stop-gradient test.}
To test whether these high-frequency gradients \emph{cause} the wide-band collapse, we train $3^{11}$ RayRoPE with its forward pass untouched while blocking only the depth gradient from the six frequencies above one rotation per patch (\cref{fig:mechanism}(c), each intervention paired with its matched $3^{11}$ run; supplement).
This backward-only change recovers 93\% of the PSNR gap to $243\times$ in the first seed ($24.54 \to 26.47\dB$; \cref{tab:official}) and 90\% in the second ($25.31 \to 26.52\dB$), keeping the two stopped seeds within $0.06\dB$ (vs.\ $1.85\dB$ spread across unstopped $3^{11}$ runs) and improving depth correlation with triangulated 3D points.
Removing those six rotations altogether recovers a similar 92\%, whereas halving all depth gradients recovers 44\% and stopping the six \emph{low} frequencies drops PSNR to $21.74\dB$.
With 16-pixel patches, stopping the same six frequencies recovers 74\% (supplement).
Computing camera geometry in single precision narrows the spread of three seeds to $0.23\dB$ but closes only a third of the gap (35\% and 31\% in two seeds), leaving a $0.9\dB$ drop, whereas applying the stop-gradient atop single-precision geometry adds $+0.70$ and $+0.75\dB$, closing 77\% and 81\% of that remaining gap (85\% and 87\% overall; supplement).
Because stop-gradient modifies only the backward pass---leaving the forward operator identical for any given weights---this identifies the backward high-frequency depth-gradient path as the primary driver of wide-band collapse.

\paragraph{Window smoothing.}
Replacing RayRoPE's sharp interval with a learned-width Gaussian recovers only 27\% of the gap (\cref{fig:mechanism}(c)), even though predicted widths do not collapse (supplement), so window smoothness alone does not explain the full gap.
Adding a fixed query-plane Gaussian floor ($\sigma_0^2 = 1/12$ patch$^2$) to RayRoPE recovers 72\% with an interval and 79\% with a Gaussian without stopping gradients, yet trails $243\times$ by $0.3$--$0.4\dB$ (vs.\ $0.1\dB$ when dropping the six high frequencies) because an unconditional one-patch floor also damps useful frequencies near one rotation per patch---including within the same view.

\paragraph{Same-view smoothing in WorldRoPE.}
WorldRoPE exhibits this exact trade-off through the projected 3D footprint of each source patch.
Across distinct views this footprint varies with depth and pose, but within the same view depth cancels out and it reduces to a constant $1/12$ patch$^2$ variance, muting upper image-plane frequencies even though same-view positions do not depend on depth.
Omitting the footprint leaves the default band unchanged, whereas under $3^{11}$ it recovers $0.33\dB$ at 25M ($0.24\dB$ when applied cross-view only) and $+0.20\dB$ at 47M---improving 47M WorldRoPE by $+0.14$ and $+0.17\dB$ over its default in two seeds as cross-view pairs access finer frequencies while same-view channels keep their full amplitude (supplement).

\section{Discussion}
\label{sec:discussion}

\paragraph{Practical guidelines.}
We make four recommendations:
\begin{enumerate}[leftmargin=*,itemsep=1pt,topsep=2pt]
  \item \textbf{Set bands directly in image coordinates, not via $d$.}
  Specify $\omega_0$ and $\omega_{n-1}$ in image units rather than via $d$ or a fixed 2D grid schedule; on $256\times256$ RealEstate10K (8-px patches), top frequencies of $0.3$--$3$ rotations/patch ($10$--$90$/image) work reliably across all three encodings.
  \item \textbf{Keep and populate the span in small heads.}
  When small heads leave few channels per coordinate, preserve the target range and populate it with intermediate frequencies (e.g., one ray with four frequencies across $243\times$ at $d=48$, rather than truncating to $3\times$ or stretching two frequencies; supplement).
  \item \textbf{Avoid unconditional same-view damping.}
  Within a view, projected image coordinates do not depend on depth, so fixed floors or projected patch footprints can unnecessarily damp wide-band channels there (\cref{sec:verification}).
  \item \textbf{Report and match bands in benchmarks.}
  State rotated coordinates, frequency counts, and ranges alongside $d$, and compare encodings at matched bands or across head sizes (\cref{fig:sweep}(a)) before attributing residual gaps to the geometric formulation.
\end{enumerate}

\paragraph{Limitations.}
Our experiments focus on feed-forward view synthesis and posed two-view depth at moderate model scales; in a 24-layer (95M) model at $d=48$, widening RayRoPE to $243\times$ still gains $+0.4\dB$ (vs.\ $+0.9\dB$ at 25M), narrowing the gap to WorldRoPE from $0.7$ to $0.3\dB$ as depth partially compensates (supplement).
Patch-size sweeps favor fixing frequencies per image rather than per patch (\cref{fig:sweep}(c)), whereas exact scaling at higher resolutions, video, and $\omega_0$ remains open.
Secondary wide-band ablations (floor, 25M footprint, and patch-size sweeps) use single-seed mixed-precision runs (\cref{sec:verification}; supplement).
Finally, our gradient analysis bounds single- and pair-coefficient derivatives rather than full training dynamics, leaving secondary differences---such as why learned-width Gaussians tolerate $3^{11}$ better in WorldRoPE than in RayRoPE (supplement)---for future study.

\section{Conclusion}

Existing geometric RoPE implementations set frequency bands by inherited default rules, and cross-method comparisons rarely match or report them---even though the band strongly governs accuracy and can invert benchmark rankings.
Overly narrow bands blur spatial correspondence across all tested encodings, while overly wide bands degrade RayRoPE because averaging over a depth interval damps high frequencies forward while leaving their depth gradients undamped.
Setting and reporting bands explicitly in image coordinates turns an overlooked confounder into a principled design choice for multi-view transformers.

\clearpage
{\small
\bibliographystyle{ieeenat_fullname}
\bibliography{references_verified,references_extra}
}

\maketitlesupplementary

\setcounter{section}{0}\setcounter{figure}{0}\setcounter{table}{0}\setcounter{equation}{0}
\renewcommand{\thesection}{\Alph{section}}
\renewcommand{\thefigure}{S\arabic{figure}}
\renewcommand{\thetable}{S\arabic{table}}
\renewcommand{\theequation}{S\arabic{equation}}
\def\theHsection{S\arabic{section}}
\def\theHfigure{S\arabic{figure}}
\def\theHtable{S\arabic{table}}
\def\theHequation{S\arabic{equation}}
\providecommand{\mref}[1]{\ref*{#1}}
\renewcommand{\dB}{\,\text{dB}\xspace}
\renewcommand{\dblfloatpagefraction}{0.75}
\setlength{\dbltextfloatsep}{10pt plus 2pt minus 2pt}
\setlength{\dblfloatsep}{6pt plus 2pt minus 2pt}
\setlength{\textfloatsep}{10pt plus 2pt minus 2pt}

This supplement provides implementation details and statistical protocols (\cref{sec:s-impl}), default frequency rules and an audit of published evaluations (\cref{sec:s-rules}), extended experiments on RealEstate10K (including zero-shot transfer to ACID; \cref{sec:s-re10k}) and ScanNet++ (\cref{sec:s-scannet}), mathematical derivations and mechanism ablations (\cref{sec:s-analysis}), full seed-level results for two-view depth estimation (\cref{sec:s-depth}), and additional qualitative comparisons (\cref{sec:s-qual}).
Section, figure, table, and equation numbers without an ``S'' prefix refer to the main paper.
An accompanying project webpage (\url{https://bispl-wm.github.io/MindTheBand/}) provides a one-minute overview video, camera-path videos, and interactive frequency-band explorers across test scenes.
Upon publication, we will release all training code, configs, and evaluation logs for the view-synthesis and two-view depth experiments.

\section{Implementation Details}
\label{sec:s-impl}

\subsection{Official WorldRoPE Codebase}

\paragraph{Training and evaluation.}
Our primary RealEstate10K experiments follow the public WorldRoPE repository~\cite{worldrope2026}: 80K mixed-precision steps, batch size 16, AdamW (learning rate $4\times10^{-4}$, \num{2500}-step warm-up, cosine decay), no gradient clipping, and MSE plus $0.5\times$ VGG perceptual loss.
Unless noted otherwise, runs share a single random seed for initialization and batch order.
The mixed-precision gradient scaler routinely backs off on overflow roughly 30--40 times per 80K run across all encodings, bands, and stop-gradient variants; this count therefore reflects standard dynamic scaling rather than training instability.
Models train on RealEstate10K at $256\times256$ with two context views and one target view, and evaluate on all \num{6474} test scenes with two context views and three target views per scene.
The 25M LVSM uses 6 layers, width 768, and 16 heads ($d=48$); the 95M model keeps the same width and heads with 24 layers; and the 47M model uses 6 layers, width 1152, and 8 heads ($d=144$).
Following the repository recipe, all three encodings concatenate camera-frame ray directions to input tokens, and 47M WorldRoPE uses gradient accumulation to fit memory without changing the batch sequence.

\paragraph{RayRoPE variants.}
The WorldRoPE repository implements RayRoPE following the released RayRoPE code, sampling three rays per patch ($C=12$ rotated coordinates), whereas the RayRoPE paper formulates one ray per patch ($C=6$).
Across all band sweeps, we hold each coordinate's lowest frequency $\omega_0$ fixed and vary only the ray count and geometric step ratio: spanning a target range $R = \omega_{n-1}/\omega_0$ across $n$ frequencies uses step ratio $R^{1/(n-1)}$.
As a control at $d=48$ (\cref{tab:s-official}), we also test a single-ray variant with two distinct frequencies repeated twice to match the exact two-frequency band of the three-ray default.
Because half-precision rotation angles overflow at $3^{11}$, we compute the angles and averaged rotation coefficients in single precision for all 47M single-ray runs, all $2187\times$ runs, and the 4-head $243\times$ run.
Unless stated otherwise (the single-precision geometry controls in \cref{tab:s-official-mods,sec:s-analysis}), camera projections and projected coordinates entering these angles remain in half precision, matching the released recipe.
WorldRoPE already computes camera projection, projected positions, and rotation angles in single precision, so its wide-band runs (\cref{tab:s-otherbands}) should be compared beside the single-precision-geometry RayRoPE rows of \cref{tab:s-official-mods}.
URoPE keeps query grid positions in single precision but projects key and value positions in half precision; for wide-band URoPE (\cref{tab:s-otherbands}), we likewise form angles in single precision from those positions.
Finally, we retain the repository's inverse-depth coordinate in official-code experiments and verify in \cref{tab:s-denc} that depth and inverse depth exhibit the same band sensitivity.

\paragraph{Head-count sweep.}
For Fig.~\mref{fig:sweep}(a), we train the 25M architecture (width 768) with 16, 8, and 4 heads ($d=48, 96, 192$), keeping four depth anchors for URoPE and four mixture components for WorldRoPE at every head count.

\paragraph{Released checkpoints and reproducibility.}
Our retrained 25M WorldRoPE and URoPE models match the authors' released checkpoints on the test split (\cref{tab:s-official}).
Evaluated with two flags restored to match its original training setup (depth instead of inverse depth, and no camera-axis flip in the pose adapter), the released 47M RayRoPE checkpoint reaches $24.71\dB$.
For 25M ($d=48$) RayRoPE, the WorldRoPE repository ships neither a pretrained checkpoint nor a separate frequency config; training its released three-ray implementation under the identical 25M recipe yields $25.55\dB$ ($25.57\dB$ on a second seed) versus $26.17\dB$ reported in~\cite{worldrope2026}.
Using depth instead of inverse depth gives the same $25.55\dB$, and we cannot trace the discrepancy further without the original config.
All comparisons in our tables therefore come from models trained under a uniform protocol.
The fixed-width head-count sweep (Fig.~\mref{fig:sweep}(a)) isolates the band directly within that single codebase: widening the 25M model to four frequencies over $243\times$ raises PSNR from $25.55$ to $26.44\dB$, while changing ray count alone has no effect (\cref{tab:s-official}).

\subsection{Unified Codebase}
\label{sec:s-code}

\paragraph{Architecture and training.}
Our ScanNet++ benchmarks, seven-factor design study, and exploratory ablations in \cref{sec:s-analysis} use our unified 47M LVSM codebase (6 decoder-only layers~\cite{jin2025lvsm}, width 1152, 8 heads of $d=144$).
Unless noted otherwise, models train for 20K steps (batch size 16, AdamW at $4\times10^{-4}$ with warm-up and cosine decay, gradient clipping, MSE plus VGG loss).
Although the backbone uses bfloat16 mixed precision, all RayRoPE runs here compute camera geometry, projected coordinates, rotation angles, and averaged coefficients in single precision and apply them to single-precision features, avoiding the half-precision coordinate rounding seen in the official code (\cref{sec:s-analysis}).
Within each seed, all methods share initial weights and training batches.

\paragraph{ScanNet++ and baseline implementation details.}
We use ScanNet++ v2 DSLR imagery~\cite{yeshwanth2023scannetpp} with 822 training scenes and 50 validation scenes, where each frame is available as both original fisheye and undistorted pinhole at $256\times256$.
Design settings were chosen on 32 scenes held out from the training split rather than on the 50 validation scenes, which serve as the test set reported in Tab.~\mref{tab:scannet}.
During training, each view is sampled independently as fisheye or pinhole with probability $0.5$ (two context views, one intermediate target view).
Unless noted otherwise, we report valid-pixel PSNR on fisheye inputs and targets averaged over three seeds.
Pinhole-defined encodings are generalized to fisheye cameras so they reduce exactly to their original form on pinhole views, with three implementation notes:
\begin{itemize}[leftmargin=*,itemsep=1pt,topsep=2pt]
  \item \textbf{Pl\"ucker + 2D RoPE and RoPE-Mixed.}
  Following recent multi-view benchmarks~\cite{li2025prope,xie2026urope,worldrope2026}, our ``Pl\"ucker + 2D RoPE'' baseline pairs world-frame Pl\"ucker rays~\cite{jin2025lvsm} with 2D axial patch RoPE~\cite{heo2024rope} (or learned-frequency RoPE-Mixed~\cite{heo2024rope} in \cref{tab:s-methods}).
  All other encodings take camera-frame ray directions at the input, except Wei et al.~\cite{hetero2026}, which uses its patch-local ray map.
  \item \textbf{Wei et al.~\cite{hetero2026}.}
  Because official code is unavailable, we reimplement the method atop UCPE~\cite{ucpe2025}.
  Although Eq.~3 of~\cite{hetero2026} writes the transform on queries and keys only, UCPE's released operator also transforms values and outputs; we retain those value/output transforms ($+0.33\dB$ in preliminary checks) and scale camera-local ray angles by 16 patch units per radian (matching ScanNet++'s fisheye focal length in 8-pixel patches, giving $\omega_{n-1}=0.093$ rotations/patch on RealEstate10K).
  \item \textbf{G-ray~\cite{gray2026} and WorldRoPE$^\ast$~\cite{worldrope2026}.}
  We evaluate G-ray's RayRoPE-hosted variant (which uses RayRoPE's $3^k$ progression on camera-local ray angles).
  WorldRoPE$^\ast$, implemented prior to the official WorldRoPE code release, rotates image-plane coordinates only using URoPE's top-down rule (\cref{tab:s-rules}: 36 frequencies over $88\times$ at $d=144$) and parameterizes Gaussians on ScanNet++ by ray distance rather than optical-axis depth.
\end{itemize}

\begin{table}[t]
  \centering
  \caption{\textbf{Depth vs.\ inverse depth as RayRoPE's depth coordinate} in the 47M model across both codebases (RealEstate10K test PSNR at 80K steps; machine B for official $3^{11}$ inverse depth, \cref{tab:s-official-seeds} lists all runs).}
  \label{tab:s-denc}
  \footnotesize
  \setlength{\tabcolsep}{5pt}
  \begin{tabular}{@{}ccrcccc@{}}
    \toprule
    & & & \multicolumn{2}{c}{Unified codebase} & \multicolumn{2}{c}{Official code} \\
    \cmidrule(lr){4-5}\cmidrule(l){6-7}
    Rays & Freq. & Range & Depth & Inverse & Depth & Inverse \\
    \midrule
    3 & 6 & $243\times$ & 26.36 & 26.37 & 26.59 & 26.66 \\
    1 & 12 & $3^{11}$ & 25.32 & 25.68 & 24.90 & 25.22 \\
    1 & 12 & $243\times$ & 26.30 & 26.20 & 26.59 & 26.61 \\
    \bottomrule
  \end{tabular}
\end{table}

\paragraph{Depth coordinate.}
Under the three-ray $243\times$ default, depth and inverse depth differ by under $0.1\dB$ in both codebases (\cref{tab:s-denc}).
Under single-ray $3^{11}$, both parameterizations drop substantially relative to $243\times$: by $0.5$--$1.0\dB$ in the unified codebase and $1.4$--$2.1\dB$ in the official code (including an inverse-depth run with a mid-training loss jump; \cref{tab:s-official-seeds}).
The wide-band drop is large under both parameterizations, and somewhat larger with depth; our analysis model uses depth.

\paragraph{Analysis model.}
The mechanism screens in \cref{sec:s-analysis} use 47M single-ray RayRoPE ($C=6$, $n=12$) on RealEstate10K under an 80K schedule.
Rapid ablations (\cref{tab:s-modifications}) report mean validation PSNR from 10K to 20K steps on 256 held-out scenes, while key configurations run to 80K steps on the full test set (\cref{tab:s-trained}).
Our unified codebase reads frames pre-resized and center-cropped with a Lanczos filter rather than resizing on the fly as in the official loader, so absolute PSNRs are not directly comparable across codebases.
In addition, its 80K RealEstate10K runs (including the analysis model) receive Pl\"ucker rays at the input, whereas its 20K runs (\cref{tab:s-methods,tab:s-factors}) use camera-frame rays as on ScanNet++; high-frequency stop-gradient recovery holds under both input conventions (\cref{sec:s-re10k,sec:s-analysis}).

\subsection{Statistics}

Throughout our tables, bracketed intervals $[a, b]$ report 95\% confidence intervals from \num{5000} paired scene-bootstrap resamples, capturing scene-sampling uncertainty rather than training-seed variance.
We quantify training variance separately: rerunning the analysis model shifts 10K--20K validation PSNR by $0.03\dB$ (up to $0.06\dB$ in \cref{tab:s-patch}), while full unified-codebase runs vary across seeds by up to $\pm0.2\dB$ on ScanNet++ and up to $0.13\dB$ SD on RealEstate10K.
In the official codebase, extra seeds of the four main 25M settings shift test PSNR by at most $0.11\dB$ and keep the same ranking in every seed (WorldRoPE 26.77, 26.67; $243\times$ RayRoPE 26.44, 26.44, 26.41; URoPE 26.26, 26.34, 26.34; default RayRoPE 25.55, 25.57\dB).
Repeating default 47M WorldRoPE with the same seed across three GPU machines (A--C, identical code and data) gives $26.69$, $26.80$, and $26.72\dB$ (a $0.12\dB$ span).
This $0.12\dB$ figure is the largest stable-regime difference quoted in the paper, so gaps of ${\sim}0.1\dB$ or less between single runs lie within run-to-run variation even when scene-bootstrap intervals exclude zero.

For wide-band interventions, gap recovery is $(m - w)/(g - w)$, where $m$, $w$, and $g$ denote the modified $3^{11}$, unmodified $3^{11}$, and $243\times$ models, with bootstrap intervals resampled jointly across all three.
Each $3^{11}$ modification is paired with the matched-seed unmodified $3^{11}$ run on the same machine where available (\cref{tab:s-official-seeds,tab:s-official-mods}).
Because the $243\times$ reference moves by under $0.1\dB$ across machines, precision settings, and seeds (\cref{tab:s-official-seeds,tab:s-official-mods}), the machine-A run ($26.61\dB$) serves as the shared first-seed reference.

\section{Frequency Rules and Published Evaluations}
\label{sec:s-rules}

\paragraph{Units.}
To compare methods on a common physical axis, we express image-plane frequencies in \emph{rotations per patch}.
Patch indices are already measured in patches.
A ray angle or focal-normalized coordinate advances by $p/f$ per $p$-pixel patch ($0.036$ on $256\times256$ RealEstate10K with $p=8$), whereas the image-normalized projected ray direction in RayRoPE and CRePE advances by $1/32$ per patch near the image center.

\begin{table*}[t]
  \centering
  \caption{\textbf{Default image-plane frequency schedules of evaluated implementations.}
  Cells report frequency count / range ($\omega_{n-1}/\omega_0$) at each head dimension $d$; \emph{Highest} gives $\omega_{n-1}$ at $d=144$ on $256\times256$ RealEstate10K in rotations per 8-pixel patch.
  $^\dagger$Uses our $16$-patch/radian angle scale (\cref{sec:s-code}).}
  \label{tab:s-rules}
  \footnotesize
  \begin{tabular}{@{}lllccccr@{}}
    \toprule
    Method & Image-plane coordinate & Rule & $d{=}48$ & $96$ & $144$ & $192$ & Highest \\
    \midrule
    2D axial RoPE & patch index & highest fixed & 12 / 68 & 24 / 83 & 36 / 88 & 48 / 91 & 0.16 \\
    GTA, PRoPE, DPPE & patch index & highest fixed & 6 / 46 & 12 / 68 & 18 / 77 & 24 / 83 & 0.16 \\
    UCPE & patch index & highest fixed & 6 / 46 & 12 / 68 & 18 / 77 & 24 / 83 & 0.16 \\
    Wei et al.$^\dagger$ & ray angle & highest fixed & 6 / 46 & 12 / 68 & 18 / 77 & 24 / 83 & 0.093 \\
    SCoPE & patch index & highest fixed & 12 / 68 & 24 / 83 & 36 / 88 & 48 / 91 & 0.16 \\
    MeRoPE & patch index & highest fixed & -- & -- & 12 / 68 & 24 / 83 & 0.16 \\
    URoPE & projected patch position & highest fixed & 12 / 68 & 24 / 83 & 36 / 88 & 48 / 91 & 0.16 \\
    WorldRoPE & projected patch position & both fixed & 8 / 68 & 16 / 68 & 24 / 68 & 32 / 68 & 0.16 \\
    RayRoPE, 3 rays & projected ray direction & lowest fixed & 2 / 3 & 4 / 27 & 6 / 243 & 8 / 2187 & 0.95 \\
    G-ray (RayRoPE host) & ray angle & lowest fixed & 2 / 3 & 4 / 27 & 6 / 243 & 8 / 2187 & 0.7 \\
    CRePE & projected ray direction & lowest fixed & 2 / 3 & 4 / 27 & 6 / 243 & 8 / 2187 & 0.95 \\
    RayRoPE, 1 ray & projected ray direction & lowest fixed & 4 / 27 & 8 / 2187 & 12 / $1.8{\times}10^{5}$ & 16 / $1.4{\times}10^{7}$ & 692 \\
    \bottomrule
  \end{tabular}
\end{table*}

\paragraph{Default schedules.}
\Cref{tab:s-rules} summarizes each method's default image-plane progression.
Top-down methods inherit 2D axial ViT RoPE ($100^{-k/n}$ rad/patch~\cite{heo2024rope}), fixing the ceiling at $1$ rad ($\approx 0.16$ rotations) per patch at every head size (except RoPE-Mixed, which learns frequency vectors freely).
RayRoPE, RayRoPE-hosted G-ray ($P_{\max}=4\pi$ rad, ratio 3~\cite{gray2026}), and CRePE instead fix $\omega_0$ ($1/8$ rotation across the image for RayRoPE's direction) and triple each step (Eq.~(\mref{eq:growing})); on non-RayRoPE hosts, G-ray uses a top-down base-100 schedule ($\alpha=10$ on GTA, $\alpha=1$ in 3D reconstruction~\cite{gray2026}).
Three-ray RayRoPE rotates 12 scalars (6 direction, 3 depth, 3 camera center), whereas single-ray RayRoPE rotates 6 (2 direction, 1 depth, 3 camera center), applying the same $3^k$ progression to every channel group.

\begin{table*}[t]
  \centering
  \caption{\textbf{Backbone configurations in published evaluations of RayRoPE-family encodings} (supporting the audit in Sec.~\mref{sec:bands}).
  Head dimension is $d = \text{width}/\text{heads}$, extracted from each paper, its appendix, or its released code.
  $^\ast$Released code omits the RayRoPE baseline.
  $^\dagger$Paper writes $D/6 = 8$ six-channel blocks, giving $d=48$.}
  \label{tab:s-published}
  \footnotesize
  \begin{tabular}{@{}lllccl@{}}
    \toprule
    Paper & Task & Backbone & Width $\times$ heads & $d$ & Source \\
    \midrule
    RayRoPE~\cite{wu2026rayrope} & view synthesis & LVSM, 47M and 150M & $1152 \times 8$ & 144 & paper \\
    \midrule
    G-ray~\cite{gray2026} & view synthesis & LVSM & $1152 \times 8$ & 144 & paper (App.~A-D, C-B) \\
    POD-RoPE~\cite{podrope2026} & view synthesis & LVSM & $1152 \times 8$ & 144 & paper \\
    POD-RoPE~\cite{podrope2026} & depth estimation & UniMatch & $144 \times 1$ & 144 & paper \\
    WorldRoPE~\cite{worldrope2026} & view synthesis & LVSM, 25M & $768 \times 16$ & 48 & paper \\
    WorldRoPE~\cite{worldrope2026} & depth estimation & UniMatch & $144 \times 3$ & 48 & paper \\
    WorldRoPE~\cite{worldrope2026} & 3D geometry & VGGT, added blocks & $1024 \times 16$ & 64 & paper (App.~C.4; heads as in VGGT-1B) \\
    MeRoPE~\cite{merope2026} & video generation & Wan2.2, camera branch & $768 \times 6$ & 128 & paper \\
    CRePE~\cite{jin2026crepe} & video generation & Wan2.1, adapter & $192 \times 1$ & 192 & paper, code \\
    URoPE~\cite{xie2026urope} & view synthesis & LVSM & $768 \times 16$ & 48 & code$^\ast$ \\
    HRoPE~\cite{hrope2026} & view synthesis & LVSM, $\sim$25M & $D/6 = 8$ blocks$^\dagger$ & 48 & paper (App.~A.3) \\
    \bottomrule
  \end{tabular}
\end{table*}

\paragraph{Audit of published evaluations.}
\Cref{tab:s-published} lists the ten external evaluations audited in Sec.~\mref{sec:bands}.
Only three match the original $d=144$ head size, while four evaluate at $d=48$, where the released three-ray rule shrinks to $n=2$ frequencies spanning $3\times$ (for URoPE, the paper omits backbone dimensions and the released LVSM config uses $768\times16$ without shipping the RayRoPE baseline).
Because rotating 12 or 6 coordinates requires $d$ divisible by 24 or 12, the released rule does not even divide evenly into WorldRoPE's $d=64$ VGGT blocks or MeRoPE's $d=128$ heads, yet neither paper specifies how RayRoPE was adapted.
Across all ten settings, only G-ray and CRePE explicitly report the resulting frequency parameters.

\section{Additional Results on RealEstate10K}
\label{sec:s-re10k}

\begin{table*}[tp]
  \centering
  \caption{\textbf{Complete RealEstate10K results in the official WorldRoPE codebase.}
  $\Delta$PSNR: paired difference to WorldRoPE at the same model and head configuration; \emph{Wins}: fraction of test scenes beating WorldRoPE.
  $^\dagger$Default $3^k$ rule on one ray; ``stop-gradient'' blocks depth gradients from its six highest frequencies.
  $^\ast$Two distinct frequencies, each used twice; $C{=}3$ omits the camera center $p_0$, giving eight frequencies per coordinate.
  $^\S$Released 47M RayRoPE checkpoint with corrected evaluation flags (\cref{sec:s-impl}).
  The 47M $3^{11}$ row is the first run (machine A); \cref{tab:s-official-seeds} lists all runs of this configuration.
  Machines (\cref{sec:s-impl}): A, except the 47M single-ray rows at $3^9$, $2187\times$, $27\times$, and $3\times$ (B), the 25M three-ray and two-frequency single-ray rows at $243\times$, the three-ray $243\times$ row at $d=96$, and the 95M block (C).}
  \label{tab:s-official}
  \footnotesize
  \begin{tabular}{@{}llcrrrrrr@{}}
\toprule
Method & Rays & Freq. & Range & PSNR & SSIM & LPIPS & $\Delta$PSNR [95\% CI] & Wins \\
\midrule
\multicolumn{9}{@{}l}{\emph{47M model: width 1152, 8 heads of 144 channels}} \\
WorldRoPE & -- & 24 & $68\times$ & 26.69 & 0.844 & 0.073 & -- & -- \\
URoPE & -- & 36 & $88\times$ & 26.24 & 0.834 & 0.078 & $-$0.45 [$-$0.47, $-$0.43] & 21\% \\
RayRoPE (default) & 3 & 6 & $243\times$ & 26.66 & 0.845 & 0.074 & $-$0.02 [$-$0.04, $-$0.01] & 47\% \\
RayRoPE$^\dagger$ & 1 & 12 & $3^{11}$ & 24.54 & 0.779 & 0.181 & $-$2.15 [$-$2.18, $-$2.12] & 3\% \\
RayRoPE & 1 & 12 & $3^{9}$ & 26.12 & 0.833 & 0.083 & $-$0.56 [$-$0.58, $-$0.55] & 18\% \\
RayRoPE & 1 & 12 & $2187\times$ & 26.52 & 0.842 & 0.076 & $-$0.17 [$-$0.18, $-$0.15] & 38\% \\
RayRoPE & 1 & 12 & $243\times$ & 26.61 & 0.844 & 0.075 & $-$0.08 [$-$0.09, $-$0.06] & 44\% \\
RayRoPE & 1 & 12 & $27\times$ & 26.38 & 0.838 & 0.076 & $-$0.31 [$-$0.32, $-$0.29] & 26\% \\
RayRoPE & 1 & 12 & $3\times$ & 25.38 & 0.812 & 0.093 & $-$1.31 [$-$1.33, $-$1.29] & 5\% \\
RayRoPE$^\dagger$, stop-gradient & 1 & 12 & $3^{11}$ & 26.47 & 0.841 & 0.077 & $-$0.22 [$-$0.24, $-$0.20] & 33\% \\
\midrule
\multicolumn{9}{@{}l}{\emph{25M model: width 768, 16 heads of 48 channels}} \\
WorldRoPE & -- & 8 & $68\times$ & 26.77 & 0.847 & 0.072 & -- & -- \\
URoPE & -- & 12 & $68\times$ & 26.26 & 0.834 & 0.079 & $-$0.52 [$-$0.54, $-$0.50] & 18\% \\
RayRoPE (default) & 3 & 2 & $3\times$ & 25.55 & 0.817 & 0.089 & $-$1.22 [$-$1.25, $-$1.20] & 6\% \\
RayRoPE & 3 & 2 & $243\times$ & 25.73 & 0.822 & 0.089 & $-$1.04 [$-$1.07, $-$1.02] & 8\% \\
RayRoPE & 1 & 2$^\ast$ & $3\times$ & 25.53 & 0.816 & 0.090 & $-$1.25 [$-$1.27, $-$1.22] & 6\% \\
RayRoPE & 1 & 2$^\ast$ & $243\times$ & 25.73 & 0.822 & 0.090 & $-$1.05 [$-$1.07, $-$1.02] & 8\% \\
RayRoPE & 1 & 4 & $3\times$ & 25.42 & 0.813 & 0.092 & $-$1.36 [$-$1.38, $-$1.33] & 5\% \\
RayRoPE$^\dagger$ & 1 & 4 & $27\times$ & 26.31 & 0.835 & 0.080 & $-$0.47 [$-$0.49, $-$0.45] & 19\% \\
RayRoPE & 1 & 4 & $68\times$ & 26.40 & 0.838 & 0.079 & $-$0.38 [$-$0.40, $-$0.36] & 23\% \\
RayRoPE & 1 & 4 & $243\times$ & 26.44 & 0.840 & 0.079 & $-$0.33 [$-$0.35, $-$0.32] & 27\% \\
RayRoPE ($C{=}3$) & 1 & 8 & $243\times$ & 26.58 & 0.843 & 0.075 & $-$0.19 [$-$0.21, $-$0.18] & 34\% \\
RayRoPE & 1 & 4 & $2187\times$ & 26.25 & 0.835 & 0.080 & $-$0.52 [$-$0.54, $-$0.50] & 18\% \\
\midrule
\multicolumn{9}{@{}l}{\emph{25M model with 8 heads of 96 channels}} \\
WorldRoPE & -- & 16 & $68\times$ & 26.60 & 0.843 & 0.075 & -- & -- \\
URoPE & -- & 24 & $83\times$ & 26.08 & 0.830 & 0.083 & $-$0.52 [$-$0.54, $-$0.50] & 17\% \\
RayRoPE (default) & 3 & 4 & $27\times$ & 26.30 & 0.836 & 0.079 & $-$0.30 [$-$0.31, $-$0.28] & 27\% \\
RayRoPE & 3 & 4 & $243\times$ & 26.38 & 0.838 & 0.079 & $-$0.22 [$-$0.24, $-$0.20] & 32\% \\
RayRoPE & 1 & 8 & $68\times$ & 26.40 & 0.838 & 0.078 & $-$0.19 [$-$0.21, $-$0.18] & 33\% \\
\midrule
\multicolumn{9}{@{}l}{\emph{25M model with 4 heads of 192 channels}} \\
WorldRoPE & -- & 32 & $68\times$ & 26.36 & 0.836 & 0.080 & -- & -- \\
URoPE & -- & 48 & $91\times$ & 25.69 & 0.819 & 0.089 & $-$0.67 [$-$0.69, $-$0.65] & 13\% \\
RayRoPE (default) & 3 & 8 & $2187\times$ & 26.22 & 0.834 & 0.083 & $-$0.14 [$-$0.15, $-$0.12] & 40\% \\
RayRoPE & 3 & 8 & $243\times$ & 26.42 & 0.839 & 0.078 & +0.06 [0.04, 0.08] & 57\% \\
RayRoPE & 1 & 16 & $68\times$ & 26.25 & 0.835 & 0.080 & $-$0.11 [$-$0.13, $-$0.09] & 39\% \\
\midrule
\multicolumn{9}{@{}l}{\emph{95M: 24 layers, width 768, 16 heads, $d{=}48$}} \\
WorldRoPE & -- & 8 & $68\times$ & 27.65 & 0.865 & 0.060 & -- & -- \\
RayRoPE (default) & 3 & 2 & $3\times$ & 26.91 & 0.851 & 0.068 & $-$0.74 [$-$0.76, $-$0.72] & 13\% \\
RayRoPE & 1 & 4 & $243\times$ & 27.31 & 0.859 & 0.065 & $-$0.34 [$-$0.35, $-$0.32] & 26\% \\
\midrule
\multicolumn{9}{@{}l}{\emph{Released checkpoints on the same test set}} \\
WorldRoPE, 25M & & & & 26.75 & 0.846 & 0.073 & & \\
URoPE, 25M & & & & 26.26 & 0.834 & 0.078 & & \\
RayRoPE, 47M$^\S$ & & & & 24.71 & 0.789 & 0.110 & & \\
\bottomrule
\end{tabular}

\end{table*}

\paragraph{Full official-code results.}
\Cref{tab:s-official} expands Tab.~\mref{tab:official} with paired bootstrap intervals, win rates, and the $d=48$ ray-count controls.
Repeating the two $3\times$ frequencies twice on a single ray matches the three-ray default, ruling out ray count or sub-patch placement as the cause of the $d=48$ deficit.
Stretching only two distinct frequencies across $243\times$ likewise gives identical scores for one or three rays, gaining under $0.2\dB$ over $3\times$ and staying $0.7\dB$ below four frequencies over $243\times$.
A wide span therefore helps only when enough intermediate frequencies populate it.
Omitting the camera center $p_0$ on a single ray ($C{=}3$, mirroring the compact ScanNet++ variant in \cref{sec:s-scannet}) doubles the frequency density at $d=48$ to eight frequencies over $243\times$.
This reaches $26.58\dB$ ($26.57\dB$ on seed~2), adding $+0.14\dB$ ($[0.13, 0.15]$; $+0.13\dB$ $[0.11, 0.14]$ on seed~2) over four-frequency $243\times$ and roughly halving the remaining gap to WorldRoPE ($-0.19\dB$ vs.\ $-0.33\dB$ on seed~1; $-0.10\dB$ vs.\ $-0.23\dB$ on seed~2).

The same trend holds in the deeper 95M backbone (24 layers at $d=48$; \cref{tab:s-official}).
Widening RayRoPE from its $3\times$ default to four frequencies across $243\times$ gains $+0.40\dB$ (winning on $74\%$ of scenes, versus $+0.89\dB$ at 25M) and narrows the gap to WorldRoPE from $0.74$ to $0.34\dB$, matching the $-0.33\dB$ band-matched margin at 25M.
Deeper networks thus partially compensate for a narrow band without altering the band-matched comparison.

\begin{table}[t]
  \centering
  \caption{\textbf{Varying head count in the 25M model} (RealEstate10K test PSNR, numerical values for Fig.~\mref{fig:sweep}(a)).
  The bottom two rows hold RayRoPE's range fixed across head sizes ($^\dagger$one ray per patch at $d=48$, where three rays afford only two frequencies).}
  \label{tab:s-heads-official}
  \footnotesize
  \setlength{\tabcolsep}{5pt}
  \begin{tabular}{@{}lccc@{}}
    \toprule
    Heads $\times$ channels & $16\times48$ & $8\times96$ & $4\times192$ \\
    \midrule
    WorldRoPE & 26.77 & 26.60 & 26.36 \\
    URoPE & 26.26 & 26.08 & 25.69 \\
    RayRoPE (default) & 25.55 & 26.30 & 26.22 \\
    \quad default range & $3\times$ & $27\times$ & $2187\times$ \\
    RayRoPE, one ray, $68\times$ & 26.40 & 26.40 & 26.25 \\
    RayRoPE, $243\times$ & 26.44$^\dagger$ & 26.38 & 26.42 \\
    \bottomrule
  \end{tabular}
\end{table}

\paragraph{Head-size sweep.}
\Cref{tab:s-heads-official} gives the exact values behind Fig.~\mref{fig:sweep}(a) (confidence intervals in \cref{tab:s-official}).
Under default rules, RayRoPE trails URoPE by $0.7\dB$ at $d=48$ yet overtakes it by $0.5\dB$ at $d=192$, while WorldRoPE leads at every head size (maintaining $26.78\dB$ even with 32 heads of $d=24$).

Matching frequency ranges isolates the band from head size.
At matched ranges ($27\times$ and $2187\times$), single-ray $d=48$ matches three-ray $d=96$ and $d=192$ within $0.03\dB$.
Holding RayRoPE at $243\times$ keeps PSNR flat within $0.06\dB$ across all three head counts, improving over the $d=96$ and $d=192$ defaults.
Holding it at WorldRoPE's $68\times$ range loses far less at $d=192$ ($0.15\dB$) than URoPE or WorldRoPE ($0.4$--$0.6\dB$), placing RayRoPE consistently between the two once its band is fixed.

\paragraph{Stop-gradient in the official codebase.}
\Cref{tab:s-official-seeds} compiles repeated runs of 47M single-ray RayRoPE across three seeds and three machines (A--C).
Unmodified $3^{11}$ trains unstably: five runs span $1.85\dB$ ($23.56$--$25.41\dB$), including nearly $0.9\dB$ for seed~1 alone across machines.
This spread comes largely from mid-training loss jumps, such as seed~1 on A jumping after 40K steps to finish $0.68\dB$ below B, and seed~3 suffering two jumps; all four $243\times$ runs stay within $0.09\dB$.

Blocking depth gradients on the six highest frequencies prevents these jumps: all four stopped runs fall within a $0.16\dB$ band and recover $88$--$93\%$ of each run's gap to $243\times$ (\cref{tab:s-official-seeds}).
Even on seed~3, stop-gradient gains $+2.81\dB$ to recover $91\%$.
Further official-code controls appear in \cref{tab:s-official-mods}.

\begin{table*}[t]
  \centering
  \caption{\textbf{Repeated runs of the 47M stop-gradient experiment} in the official codebase (single-ray RayRoPE, 8-pixel patches, test PSNR at 80K steps across seeds 1--3 and machines A--C; 95\% scene-bootstrap CIs).
  Within each block, $\Delta$ and recovery are paired against that seed's $243\times$ run.
  $^\dagger$Paired against the seed-1 $243\times$ run on A ($26.61\dB$); against seed~1's $243\times$ run on C ($26.57\dB$), stop-gradient on C closes $91\%$ $[90, 92]$ of the gap.
  Cross-seed shifts (seed~2 on B minus seed~1 on A): $+0.78$ $[0.76, 0.80]$ at $3^{11}$ ($+0.10$ $[0.08, 0.11]$ vs.\ seed~1 on B), $+0.06$ $[0.04, 0.07]$ with stop-gradient, and $+0.05$ $[0.04, 0.06]$ at $243\times$.}
  \label{tab:s-official-seeds}
  \footnotesize
  \begin{tabular}{@{}cclrrr@{}}
\toprule
Seed & Machine & Configuration & PSNR & $\Delta$ to $243\times$ [95\% CI] & Recovery [95\% CI] \\
\midrule
1 & A & $3^{11}$ & 24.54 & $-$2.07 [$-$2.10, $-$2.04] & -- \\
1 & A & \quad $+$ stop depth grad.\ (6--11) & 26.47 & $-$0.14 [$-$0.16, $-$0.13] & 93\% [92, 94] \\
1 & A & $243\times$ & 26.61 & -- & -- \\
\midrule
1 & B & $3^{11}$ & 25.22 & $-$1.39 [$-$1.41, $-$1.37]$^\dagger$ & -- \\
1 & C & $3^{11}$ & 25.41 & $-$1.20 [$-$1.22, $-$1.17]$^\dagger$ & -- \\
1 & C & \quad $+$ stop depth grad.\ (6--11) & 26.47 & $-$0.14 [$-$0.16, $-$0.13]$^\dagger$ & 88\% [87, 89]$^\dagger$ \\
1 & C & $243\times$ & 26.57 & $-$0.04 [$-$0.05, $-$0.03]$^\dagger$ & -- \\
\midrule
2 & B & $3^{11}$ & 25.31 & $-$1.34 [$-$1.37, $-$1.32] & -- \\
2 & B & \quad $+$ stop depth grad.\ (6--11) & 26.52 & $-$0.14 [$-$0.15, $-$0.12] & 90\% [89, 91] \\
2 & B & $243\times$ & 26.66 & -- & -- \\
\midrule
3 & C & $3^{11}$ & 23.56 & $-$3.08 [$-$3.12, $-$3.04] & -- \\
3 & C & \quad $+$ stop depth grad.\ (6--11) & 26.37 & $-$0.27 [$-$0.29, $-$0.26] & 91\% [91, 92] \\
3 & C & $243\times$ & 26.64 & -- & -- \\
\bottomrule
\end{tabular}

\end{table*}

\begin{table*}[t]
  \centering
  \caption{\textbf{Frequency range across patch sizes and image resolutions} in the official codebase (47M single-ray RayRoPE, 80K steps, one seed; 8-pixel cells were trained on machine B except $243\times$ (A) and $3^6$ (C), 16-pixel cells on B or C, and $128\times128$ cells on C; the $3^{11}$ cell with 8-pixel patches is the first-seed run on B of \cref{tab:s-official-seeds}).
  Because $\omega_0$ is fixed in image units, an image-tied rule predicts the same optimal range across columns, whereas a patch-tied or pixel-tied rule shifts 16-pixel patches and $128\times128$ images down by $2\times$ ($\approx 3^{4.4}$).
  PSNR on $128\times128$ targets is not directly comparable to $256\times256$ columns; $\Delta$ is paired against $243\times$ within each column.}
  \label{tab:s-official-patch}
  \footnotesize
  \setlength{\tabcolsep}{3.4pt}
  \begin{tabular}{@{}lrrrrrrrrr@{}}
\toprule
 & & \multicolumn{3}{c}{8-pixel patches} & \multicolumn{3}{c}{16-pixel patches} & \multicolumn{2}{c}{$128{\times}128$ images, 8-pixel patches} \\
\cmidrule(lr){3-5}\cmidrule(lr){6-8}\cmidrule(l){9-10}
Range & Across image & Per patch & PSNR & $\Delta$ to $243\times$ [95\% CI] & Per patch & PSNR & $\Delta$ to $243\times$ [95\% CI] & PSNR & $\Delta$ to $243\times$ [95\% CI] \\
\midrule
$3^{1}$ & 0.38 & 0.01 & 25.38 & $-$1.23 [$-$1.26, $-$1.21] & -- & -- & -- & -- & -- \\
$3^{3}$ & 3.4 & 0.11 & 26.38 & $-$0.23 [$-$0.25, $-$0.22] & -- & -- & -- & -- & -- \\
$3^{4}$ & 10 & 0.32 & 26.54 & $-$0.07 [$-$0.09, $-$0.06] & -- & -- & -- & 24.72 & $-$0.05 [$-$0.06, $-$0.04] \\
$3^{4.5}$ & 18 & -- & -- & -- & 1.10 & 24.97 & $-$0.18 [$-$0.19, $-$0.17] & 24.66 & $-$0.10 [$-$0.11, $-$0.09] \\
$3^{5}$ $=243\times$ & 30 & 0.95 & 26.61 & -- & 1.90 & 25.15 & -- & 24.77 & -- \\
$3^{5.5}$ & 53 & -- & -- & -- & 3.3 & 25.15 & +0.01 [0.00, 0.02] & -- & -- \\
$3^{6}$ & 91 & 2.85 & 26.60 & $-$0.01 [$-$0.02, 0.01] & 5.7 & 25.19 & +0.04 [0.03, 0.05] & 24.80 & +0.03 [0.02, 0.04] \\
$3^{6.5}$ & 158 & -- & -- & -- & 9.9 & 25.19 & +0.04 [0.03, 0.05] & -- & -- \\
$3^{7}$ & 273 & 8.5 & 26.52 & $-$0.09 [$-$0.10, $-$0.08] & 17 & 25.10 & $-$0.04 [$-$0.05, $-$0.03] & 24.75 & $-$0.02 [$-$0.03, $-$0.01] \\
$3^{9}$ & 2{,}460 & 77 & 26.12 & $-$0.49 [$-$0.50, $-$0.47] & 154 & 24.94 & $-$0.21 [$-$0.22, $-$0.20] & -- & -- \\
$3^{11}$ & 22{,}143 & 692 & 25.22 & $-$1.39 [$-$1.41, $-$1.37] & 1{,}384 & 23.87 & $-$1.28 [$-$1.30, $-$1.26] & -- & -- \\
\bottomrule
\end{tabular}

\end{table*}

\paragraph{Frequency range across patch sizes and resolutions.}
\Cref{tab:s-official-patch} and Fig.~\mref{fig:sweep}(c) compare 8- and 16-pixel patches on $256\times256$ images.
Doubling the patch size doubles rotations per patch at every range, so a patch-tied rule would shift the plateau down by $2\times$ (near $3^{4.4}$) and worsen wide-band drops.
The measured curves do not shift downward: with 8-pixel patches, $3^4$--$3^7$ stay within $0.09\dB$; with 16-pixel patches, $3^{4.5}$ already trails $243\times$ by $0.18\dB$ while $3^5$--$3^7$ stay within $0.09\dB$, and wide-band losses at 16 pixels are no larger than at 8 pixels, matching the analysis model (\cref{tab:s-patch}).
Halving the image resolution to $128\times128$ likewise shows no downward shift, keeping all five tested ranges ($3^4$--$3^7$) within $0.14\dB$.

On 16-pixel $3^{11}$, stopping high-frequency depth gradients still recovers $74\%$ of the loss, while single-precision geometry alone recovers $40\%$ (single-seed controls).
These sweeps favor aligning bands in image coordinates rather than patch units, though single-seed runs and tokenization changes leave an exact scaling law open.

\begin{table}[t]
  \centering
  \caption{\textbf{URoPE and WorldRoPE under RayRoPE's image-plane bands} (official codebase, test PSNR; $\omega_0 = 1/8$ rotation/image for non-default columns).
  25M cells ran on machine A except $77\times$ and $768\times$ (C); superscripts mark machines for 47M cells ($^\dagger$C to 13K steps, then B).
  Same-machine 47M controls with footprint: $26.82$ and $26.84\dB$ (A) at $256\times$ and $768\times$; default WorldRoPE is $26.80\dB$ (B) and $26.72\dB$ (C).
  Other cross-machine reruns: 47M URoPE $3^{11}$ is $26.22\dB$ (A) and $26.33\dB$ (C, fp32 positions).}
  \label{tab:s-otherbands}
  \footnotesize
  \setlength{\tabcolsep}{2.7pt}
  \begin{tabular}{@{}lcccccc@{}}
    \toprule
    Range & default & $3\times$ & $77\times$ & $256\times$ & $768\times$ & $3^{11}$ \\
    Highest (rot./patch) & 0.16 & 0.012 & 0.3 & 1 & 3 & 692 \\
    \midrule
    \multicolumn{7}{@{}l}{\emph{(a) Band sweep (25M: $16\times48$; 47M: $8\times144$)}} \\
    URoPE (25M) & 26.26 & 25.67 & 26.36 & 26.32 & 26.27 & 26.11 \\
    WorldRoPE (25M) & 26.77 & 25.75 & 26.67 & 26.79 & 26.65 & 26.46 \\
    URoPE (47M) & 26.24$^{\text{A}}$ & 25.50$^{\text{B}}$ & -- & 26.23$^{\text{B}}$ & 26.27$^{\text{B}}$ & 26.18$^{\text{C}}$ \\
    WorldRoPE (47M) & 26.69$^{\text{A}}$ & 25.51$^{\text{B}}$ & -- & 26.78$^{\text{B}}$ & 26.84$^{\text{B}}$ & 26.75$^{\dagger}$ \\
    \bottomrule
  \end{tabular}

  \vspace{5pt}
  \setlength{\tabcolsep}{2.8pt}
  \begin{tabular}{@{}lcccc@{}}
    \toprule
    \emph{(b) Targeted controls} & default & $256\times$ & $768\times$ & $3^{11}$ \\
    \midrule
    WorldRoPE (25M), no patch footprint & 26.76 & -- & -- & 26.79 \\
    WorldRoPE (25M), cross-view only & -- & -- & -- & 26.70 \\
    URoPE (25M), fp32 positions & -- & -- & -- & 26.14 \\
    WorldRoPE (47M), no patch footprint & 26.81$^{\text{C}}$ & 27.02$^{\text{A}}$ & 27.03$^{\text{A}}$ & 26.95$^{\text{B}}$ \\
    WorldRoPE (47M), cross-view only & -- & 26.93$^{\text{A}}$ & -- & 26.77$^{\text{B}}$ \\
    URoPE (47M), fp32 positions & -- & -- & -- & 26.31$^{\text{A}}$ \\
    \bottomrule
  \end{tabular}
\end{table}

\begin{figure*}[t]
  \centering
  \includegraphics[width=\textwidth]{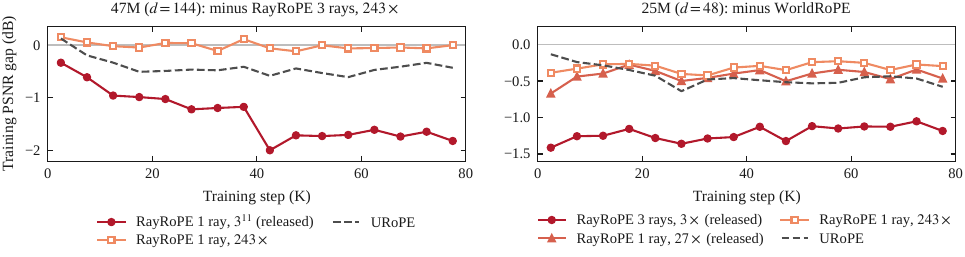}
  \caption{\textbf{Training PSNR differences over optimization} in the official codebase (averaged in 5K-step bins; ``released'' in the legend denotes the default $3^k$ rule), relative to three-ray RayRoPE ($243\times$) at 47M (\emph{left}) and WorldRoPE at 25M (\emph{right}).}
  \label{fig:s-train}
\end{figure*}

\paragraph{WorldRoPE under wide bands and its source-patch footprint.}
To separate universal band effects from RayRoPE-specific ones, \cref{tab:s-otherbands} sweeps URoPE and WorldRoPE across five RayRoPE-like image-plane bands while leaving depth frequencies untouched.
Recall that WorldRoPE builds each 3D Gaussian from a fixed source-patch covariance $\mathrm{diag}(\Delta u^2/12, \Delta v^2/12)$ back-projected at predicted depth plus along-ray variance, then splats it into the query view~\cite{worldrope2026}.
Across views, this footprint depends on depth and pose; within the \emph{same} view, depth cancels out and the covariance reduces identically to $\mathrm{diag}(1/12, 1/12)$ patch$^2$.

At $3^{11}$, this source-patch footprint lowers PSNR rather than improving it.
At 25M, $3^{11}$ drops WorldRoPE by $0.32\dB$ with the footprint enabled, whereas omitting it recovers the full loss ($+0.33\dB$, matching the default; \cref{tab:s-otherbands}).
At the default band, omitting the footprint changes 25M PSNR by only $-0.02\dB$, so the footprint does not account for WorldRoPE's lead at 25M.

Probing the trained 25M $3^{11}$ models shows the cause.
With the footprint enabled, all four frequencies above $1$ rotation/patch are damped near zero on every pair, even within the same view, leaving only half the frequencies active.
Without the footprint, learned cross-view widths stay moderate (retaining active cross-view amplitudes), while same-view projections land unattenuated on integer patch centers.
In \cref{fig:s-analysis}(c), zeroing those four frequencies within the same view drops PSNR by $5.4\dB$ versus only $0.1\dB$ across views, and applying the footprint \emph{only} across distinct views recovers three quarters of the loss ($+0.24\dB$).
Learned-width Gaussians alone therefore suffice for WorldRoPE under an image-plane $3^{11}$ band, unlike RayRoPE where a learned Gaussian closes only $27\%$ (\cref{tab:s-official-mods}; probed directly in \cref{sec:s-analysis}); note that wide WorldRoPE widens only the image-plane coordinates (keeping depth at $2$ rad/depth) and also differs in mixture weighting and width parameterization.

At 47M, $3^{11}$ costs WorldRoPE at most $0.05\dB$ with the footprint, and omitting it adds $+0.20\dB$ (winning on $70\%$ of scenes) to exceed the matched default by $+0.14\dB$; a second seed confirms this gain ($26.96\dB$, $+0.17\dB$ over its default on C).
Omitting the footprint gives a similar $+0.19$ to $+0.20\dB$ gain at $256\times$ and $768\times$ against the same-machine controls noted in \cref{tab:s-otherbands}.
Applying the footprint only across distinct views at 47M separates the same-view and cross-view effects (\cref{tab:s-otherbands}).
At $256\times$ (where all 24 frequencies lie below $1$ rotation/patch and the top bands are heavily damped within the same view), removing same-view damping alone reaches $26.93\dB$ ($+0.11\dB$ $[0.09, 0.12]$ over $26.82\dB$, or $54\%$ $[49, 60]$ of the $+0.20\dB$ footprint-free gain), with the remaining $+0.09\dB$ ($[0.08, 0.10]$) coming from cross-view pairs.
At $3^{11}$, where 47M already retains 11 image-plane frequencies below $1$ rotation/patch, removing same-view damping alone gains only $+0.02\dB$ ($[0.00, 0.03]$, $26.77\dB$), so most of the $+0.20\dB$ gain there comes from letting cross-view pairs access finer sub-patch frequencies.
At the narrow default band, where the footprint damps amplitudes by only $4\%$, omitting it stays within machine-to-machine spread, so we claim a footprint benefit only at wider bands.

\begin{table*}[tp]
  \centering
  \caption{\textbf{Zero-shot evaluation on ACID~\cite{liu2021infinite}} (\num{1595} test scenes, evaluated with the RealEstate10K-trained official-code checkpoints of \cref{tab:s-official,tab:s-official-seeds} without fine-tuning).
  $\Delta$PSNR and \emph{Wins} are paired against the first row of each block ($8\times96$ and $4\times192$ sub-blocks paired against their respective WorldRoPE row; 95\% scene-bootstrap CIs).
  $^\dagger$Default $3^k$ rule on one ray.
  Additional seeds at $16\times48$ confirm the same ranking: WorldRoPE $28.56\dB$ (seed~2), URoPE $28.42$ and $28.39\dB$ (seeds~2--3), default $3\times$ RayRoPE $27.65\dB$ (seed~2), and $243\times$ single-ray RayRoPE $28.26$ and $28.19\dB$ (seeds~2--3); the jumped machine-A 47M $3^{11}$ run scores $26.59\dB$ (SSIM $0.765$, LPIPS $0.221$).}
  \label{tab:s-acid}
  \footnotesize
  \renewcommand{\arraystretch}{0.94}
  \begin{tabular}{@{}llcrrrrrr@{}}
\toprule
Method & Rays & Freq. & Range & PSNR & SSIM & LPIPS & $\Delta$PSNR [95\% CI] & Wins \\
\midrule
\multicolumn{9}{@{}l}{\emph{25M model: width 768, 16 heads of 48 channels}} \\
WorldRoPE & -- & 8 & $68\times$ & 28.62 & 0.835 & 0.090 & -- & -- \\
URoPE & -- & 12 & $68\times$ & 28.26 & 0.825 & 0.095 & $-$0.36 [$-$0.39, $-$0.33] & 19\% \\
RayRoPE (default) & 3 & 2 & $3\times$ & 27.64 & 0.807 & 0.106 & $-$0.98 [$-$1.03, $-$0.93] & 8\% \\
RayRoPE$^\dagger$ & 1 & 4 & $27\times$ & 28.19 & 0.823 & 0.101 & $-$0.43 [$-$0.47, $-$0.39] & 16\% \\
RayRoPE & 1 & 4 & $243\times$ & 28.24 & 0.825 & 0.100 & $-$0.39 [$-$0.43, $-$0.34] & 18\% \\
RayRoPE & 1 & 4 & $2{,}187\times$ & 28.08 & 0.821 & 0.098 & $-$0.54 [$-$0.58, $-$0.50] & 12\% \\
\midrule
\multicolumn{9}{@{}l}{\emph{25M head-size sweep (width 768): $8\times96$ and $4\times192$}} \\
WorldRoPE ($8\times96$) & -- & 16 & $68\times$ & 28.46 & 0.831 & 0.093 & -- & -- \\
URoPE ($8\times96$) & -- & 24 & $83\times$ & 28.11 & 0.819 & 0.099 & $-$0.35 [$-$0.39, $-$0.32] & 20\% \\
RayRoPE (default, $8\times96$) & 3 & 4 & $27\times$ & 28.23 & 0.824 & 0.097 & $-$0.24 [$-$0.27, $-$0.20] & 24\% \\
\addlinespace[3pt]
WorldRoPE ($4\times192$) & -- & 32 & $68\times$ & 28.20 & 0.822 & 0.100 & -- & -- \\
URoPE ($4\times192$) & -- & 48 & $91\times$ & 27.75 & 0.809 & 0.106 & $-$0.45 [$-$0.49, $-$0.42] & 14\% \\
RayRoPE (default, $4\times192$) & 3 & 8 & $2{,}187\times$ & 28.09 & 0.822 & 0.102 & $-$0.11 [$-$0.16, $-$0.07] & 37\% \\
RayRoPE ($4\times192$) & 3 & 8 & $243\times$ & 28.28 & 0.827 & 0.096 & +0.08 [0.04, 0.13] & 54\% \\
\midrule
\multicolumn{9}{@{}l}{\emph{47M model: width 1152, 8 heads of 144 channels (seed~1)}} \\
WorldRoPE & -- & 24 & $68\times$ & 28.57 & 0.833 & 0.090 & -- & -- \\
URoPE & -- & 36 & $88\times$ & 28.25 & 0.824 & 0.093 & $-$0.32 [$-$0.35, $-$0.29] & 21\% \\
RayRoPE (default) & 3 & 6 & $243\times$ & 28.48 & 0.833 & 0.092 & $-$0.08 [$-$0.12, $-$0.04] & 37\% \\
RayRoPE & 1 & 12 & $3\times$ & 27.50 & 0.801 & 0.109 & $-$1.07 [$-$1.12, $-$1.02] & 8\% \\
RayRoPE & 1 & 12 & $27\times$ & 28.33 & 0.828 & 0.092 & $-$0.24 [$-$0.28, $-$0.20] & 24\% \\
RayRoPE & 1 & 12 & $243\times$ & 28.42 & 0.831 & 0.094 & $-$0.14 [$-$0.18, $-$0.10] & 30\% \\
RayRoPE & 1 & 12 & $2{,}187\times$ & 28.33 & 0.830 & 0.094 & $-$0.23 [$-$0.27, $-$0.19] & 27\% \\
RayRoPE & 1 & 12 & $3^{9}$ & 28.00 & 0.819 & 0.103 & $-$0.57 [$-$0.62, $-$0.52] & 14\% \\
RayRoPE$^\dagger$ (B, non-jumped) & 1 & 12 & $3^{11}$ & 27.18 & 0.792 & 0.120 & $-$1.39 [$-$1.45, $-$1.33] & 6\% \\
RayRoPE$^\dagger$, stop-gradient & 1 & 12 & $3^{11}$ & 28.30 & 0.828 & 0.094 & $-$0.27 [$-$0.31, $-$0.22] & 26\% \\
\midrule
\multicolumn{9}{@{}l}{\emph{47M stop-gradient confirmation (seed~2, machine~B)}} \\
RayRoPE & 1 & 12 & $243\times$ & 28.51 & 0.834 & 0.092 & -- & -- \\
RayRoPE$^\dagger$ & 1 & 12 & $3^{11}$ & 27.38 & 0.797 & 0.162 & $-$1.13 [$-$1.17, $-$1.08] & 6\% \\
RayRoPE$^\dagger$, stop-gradient & 1 & 12 & $3^{11}$ & 28.40 & 0.831 & 0.093 & $-$0.11 [$-$0.13, $-$0.09] & 37\% \\
\bottomrule
\end{tabular}

\end{table*}

\begin{table*}[t]
  \centering
  \caption{\textbf{Default-setting comparison across methods} in our unified 47M codebase (20K steps, mean $\pm$ SD over three seeds; five seeds for PRoPE and Pl\"ucker + 2D RoPE on RealEstate10K).
  ScanNet++ columns vary context$\to$target camera models (F: fisheye, P: pinhole).
  $^\ast$Image-plane coordinates only; baseline implementation choices in \cref{sec:s-code}.}
  \label{tab:s-methods}
  \footnotesize
  \begin{tabular}{@{}lccccc@{}}
    \toprule
    & \multicolumn{4}{c}{ScanNet++ (context$\to$target)} & RealEstate10K \\
    \cmidrule(lr){2-5}\cmidrule(l){6-6}
    Method & F$\to$F & P$\to$P & P$\to$F & F$\to$P & P$\to$P \\
    \midrule
    \multicolumn{6}{@{}l}{\emph{Points on the ray at a predicted depth}} \\
    RayRoPE~\cite{wu2026rayrope} & 23.76 $\pm$ 0.09 & 23.51 $\pm$ 0.09 & 23.10 $\pm$ 0.07 & 23.84 $\pm$ 0.09 & 25.34 $\pm$ 0.09 \\
    G-ray (RayRoPE)~\cite{gray2026} & 23.81 $\pm$ 0.06 & 23.56 $\pm$ 0.06 & 23.13 $\pm$ 0.05 & 23.91 $\pm$ 0.05 & 25.26 $\pm$ 0.06 \\
    CRePE~\cite{jin2026crepe} & 23.84 $\pm$ 0.03 & 23.57 $\pm$ 0.04 & 23.15 $\pm$ 0.03 & 23.90 $\pm$ 0.03 & 25.34 $\pm$ 0.08 \\
    \multicolumn{6}{@{}l}{\emph{Projection at depth anchors / mixture}} \\
    WorldRoPE$^\ast$~\cite{worldrope2026} & 23.25 $\pm$ 0.05 & 22.96 $\pm$ 0.05 & 22.62 $\pm$ 0.06 & 23.29 $\pm$ 0.04 & 24.82 $\pm$ 0.03 \\
    URoPE~\cite{xie2026urope} & 23.16 $\pm$ 0.06 & 22.89 $\pm$ 0.04 & 22.55 $\pm$ 0.04 & 23.20 $\pm$ 0.06 & 24.61 $\pm$ 0.06 \\
    \multicolumn{6}{@{}l}{\emph{Camera transformations}} \\
    UCPE~\cite{ucpe2025} & 22.22 $\pm$ 0.05 & 22.05 $\pm$ 0.02 & 21.72 $\pm$ 0.04 & 22.29 $\pm$ 0.03 & 23.70 $\pm$ 0.11 \\
    Wei et al.~\cite{hetero2026} & 22.08 $\pm$ 0.02 & 21.92 $\pm$ 0.04 & 21.61 $\pm$ 0.02 & 22.17 $\pm$ 0.04 & 23.67 $\pm$ 0.04 \\
    GTA~\cite{miyato2024gta} & 21.35 $\pm$ 0.13 & 21.24 $\pm$ 0.10 & 20.89 $\pm$ 0.10 & 21.39 $\pm$ 0.10 & 23.59 $\pm$ 0.10 \\
    PRoPE~\cite{li2025prope} & 21.25 $\pm$ 0.19 & 21.16 $\pm$ 0.19 & 20.85 $\pm$ 0.19 & 21.32 $\pm$ 0.20 & 23.59 $\pm$ 0.09 \\
    \multicolumn{6}{@{}l}{\emph{No camera encoding in attention (Pl\"ucker inputs)}} \\
    Pl\"ucker~\cite{jin2025lvsm} + 2D RoPE & 21.39 $\pm$ 0.06 & 21.21 $\pm$ 0.03 & 20.93 $\pm$ 0.03 & 21.46 $\pm$ 0.07 & 23.07 $\pm$ 0.05 \\
    Pl\"ucker~\cite{jin2025lvsm} + RoPE-Mixed~\cite{heo2024rope} & 21.35 $\pm$ 0.12 & 21.16 $\pm$ 0.11 & 20.87 $\pm$ 0.11 & 21.43 $\pm$ 0.12 & 23.04 $\pm$ 0.06 \\
    \bottomrule
  \end{tabular}
\end{table*}

\begin{table*}[t]
  \centering
  \caption{\textbf{Main effects of seven design choices} in a 16-run fractional factorial study (numerical values for Fig.~\mref{fig:effects}; mean $\pm$ SE over three seeds when switching from the first to the second option).}
  \label{tab:s-factors}
  \footnotesize
  \setlength{\tabcolsep}{4pt}
  \begin{tabular}{@{}lllrrr@{}}
    \toprule
    Design choice & First option & Second option & ScanNet++ F$\to$F & ScanNet++ P$\to$P & RealEstate10K \\
    \midrule
    Where the rotation is applied & queries and keys & also values and outputs & $+0.34 \pm 0.06$ & $+0.33 \pm 0.05$ & $+0.37 \pm 0.02$ \\
    Image coordinates & patch index & ray angle & $+0.01 \pm 0.07$ & $+0.01 \pm 0.06$ & $-0.02 \pm 0.02$ \\
    Camera input & camera rays & local ray map~\cite{hetero2026} & $-0.15 \pm 0.04$ & $-0.16 \pm 0.04$ & $-0.11 \pm 0.02$ \\
    Relation between views & camera transformation & reprojection at two depths & $+1.86 \pm 0.05$ & $+1.70 \pm 0.03$ & $+0.99 \pm 0.03$ \\
    Channels for the cross-view part & 36 of 144 & 72 of 144 & $+0.02 \pm 0.04$ & $+0.02 \pm 0.04$ & $-0.01 \pm 0.04$ \\
    Highest patch frequency & 0.16 rotation per patch & 0.64 rotation per patch & $+0.07 \pm 0.02$ & $+0.07 \pm 0.03$ & $+0.14 \pm 0.01$ \\
    Direction of the frequencies & along the image axes & learned & $+0.03 \pm 0.05$ & $+0.03 \pm 0.04$ & $-0.05 \pm 0.01$ \\
    \bottomrule
  \end{tabular}
\end{table*}

\paragraph{Plateau and narrow-band behavior in URoPE and WorldRoPE.}
Across the moderate plateau ($0.16$--$3$ rotations/patch), both URoPE and WorldRoPE stay stable at 25M and 47M (\cref{tab:s-otherbands}).
At 25M, raising the ceiling to $0.3$, $1$, or $3$ rotations/patch stays within $0.02\dB$ of each method's default seed range (\cref{sec:s-impl}); at 47M, all moderate bands fall within run-to-run variation.

At $3^{11}$, fixed-depth URoPE drops by only $0.2$--$0.3\dB$ at 25M (across two seeds) and under $0.1\dB$ at 47M, far below RayRoPE's drop ($1.2$--$3.1\dB$ across five runs).
Projecting URoPE's key and value positions in single precision closes part of that small gap at 25M and removes it at 47M (\cref{tab:s-otherbands}).
Compressing either encoding to $3\times$, however, causes large losses ($0.6$--$1.3\dB$), matching RayRoPE's narrow-band failure.

\begin{table*}[t]
  \centering
  \caption{\textbf{Head dimensions on fisheye ScanNet++} (test PSNR on the 50 validation scenes, mean $\pm$ SD over three seeds; model width 1152 at $8\times144$ and 768 otherwise).
  (a)~Ten methods at $d=144$ vs.\ $d=48$: only the RayRoPE family (whose default range shrinks from $243\times$ to $3\times$) drops sharply.
  (b)~Fixed-width sweep at $d=96$ and $192$ with RayRoPE-family range rescaled to $243\times$.
  $^\ast$Image-plane coordinates only (\cref{sec:s-code}).}
  \label{tab:s-heads}
  \footnotesize
  \begin{minipage}[t]{0.47\textwidth}
    \centering
    \setlength{\tabcolsep}{4.5pt}
    \begin{tabular}{@{}lcc@{}}
      \toprule
      \emph{(a) Default bands} & $8\times144$ & $16\times48$ \\
      RayRoPE family & 6, $243\times$ & 2, $3\times$ \\
      \midrule
      RayRoPE & 23.76 $\pm$ 0.09 & 22.40 $\pm$ 0.04 \\
      G-ray (RayRoPE) & 23.81 $\pm$ 0.06 & 22.48 $\pm$ 0.01 \\
      CRePE & 23.84 $\pm$ 0.03 & 22.45 $\pm$ 0.08 \\
      WorldRoPE$^\ast$ & 23.25 $\pm$ 0.05 & 23.47 $\pm$ 0.06 \\
      URoPE & 23.16 $\pm$ 0.06 & 23.27 $\pm$ 0.03 \\
      UCPE & 22.22 $\pm$ 0.05 & 22.26 $\pm$ 0.05 \\
      Wei et al. & 22.08 $\pm$ 0.02 & 22.23 $\pm$ 0.02 \\
      GTA & 21.35 $\pm$ 0.13 & 21.69 $\pm$ 0.03 \\
      PRoPE & 21.25 $\pm$ 0.19 & 21.59 $\pm$ 0.10 \\
      Pl\"ucker + 2D RoPE & 21.39 $\pm$ 0.06 & 21.59 $\pm$ 0.02 \\
      \bottomrule
    \end{tabular}
  \end{minipage}\hfill
  \begin{minipage}[t]{0.50\textwidth}
    \centering
    \setlength{\tabcolsep}{4.5pt}
    \begin{tabular}{@{}lcc@{}}
      \toprule
      \emph{(b) Width-768 sweep \& $243\times$ rescaling} & Default & $243\times$ \\
      \midrule
      \multicolumn{3}{@{}l}{\emph{$8\times96$ (RayRoPE family: 4 freqs., default $27\times$)}} \\
      RayRoPE & 23.37 $\pm$ 0.02 & 23.54 $\pm$ 0.07 \\
      G-ray (RayRoPE) & 23.27 $\pm$ 0.09 & 23.49 $\pm$ 0.06 \\
      CRePE & 23.23 $\pm$ 0.13 & 23.47 $\pm$ 0.10 \\
      WorldRoPE$^\ast$ & 23.23 $\pm$ 0.13 & -- \\
      URoPE & 23.04 $\pm$ 0.06 & -- \\
      Pl\"ucker + 2D RoPE & 21.32 $\pm$ 0.12 & -- \\
      \midrule
      \multicolumn{3}{@{}l}{\emph{$4\times192$ (RayRoPE family: 8 freqs., default $2187\times$)}} \\
      RayRoPE & 23.48 $\pm$ 0.11 & 23.56 $\pm$ 0.02 \\
      Pl\"ucker + 2D RoPE & 20.62 $\pm$ 0.13 & -- \\
      \bottomrule
    \end{tabular}
  \end{minipage}
\end{table*}

\begin{figure*}[t]
  \centering
  \includegraphics[width=\textwidth]{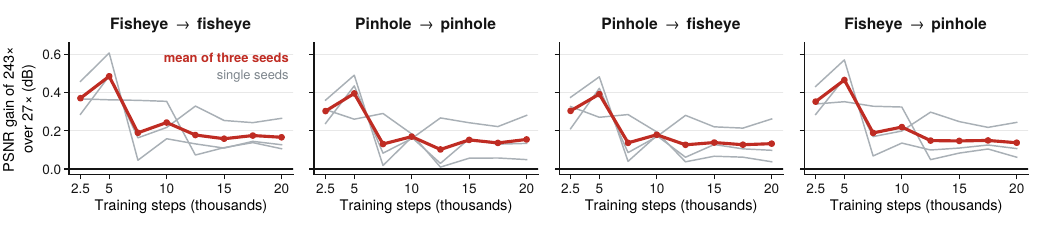}
  \caption{\textbf{Effect of widening the band across ScanNet++ camera pairs.}
  At fixed width 768 and head dimension 96, we subtract the test PSNR (50 validation scenes; \cref{sec:s-code}) of the default $27\times$ RayRoPE run from that of the $243\times$ run with the same seed. Panel titles give the context and target camera models, in that order. Each point uses 400 target views from 50 validation scenes.}
  \label{fig:s-scannetpp-band}
\end{figure*}

\begin{figure*}[p]
  \centering
  \includegraphics[width=\textwidth]{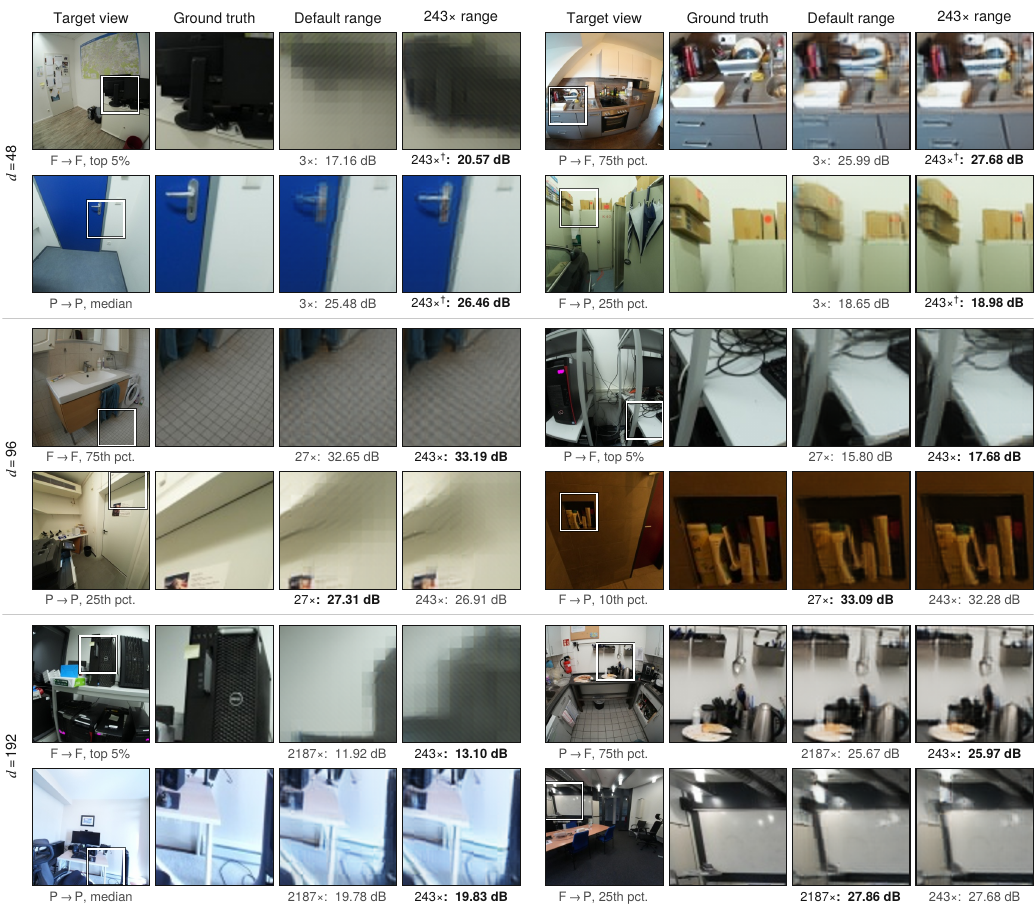}
  \caption{\textbf{Where the RayRoPE band changes ScanNet++ view synthesis.}
  Each block compares the default range of one head size with $243\times$ (width 768, seed 0, 20K steps) across all four camera pairs---homogeneous pairs on the left ($\text{F}\to\text{F}$ above, $\text{P}\to\text{P}$ below) and mixed pairs on the right ($\text{P}\to\text{F}$ above, $\text{F}\to\text{P}$ below), where F and P denote fisheye and pinhole (context $\to$ target). Targets sit at fixed quantiles of the $243\times$-minus-default gain among the 400 targets of their camera pair, shown under each target view: one from the top 5\%, the 75th percentile, the median, and the 25th percentile (at $d=96$, the 25th and 10th percentiles in the lower row). The white box marks the crop, the fully valid $80\times80$ region where the two predictions differ most; PSNR covers the full valid target, with the higher value in bold. At $d=96$ and $192$ only the band changes; the $243\times$ model at $d=48$ ($\dagger$) is the compact variant with one ray, no camera-center coordinates and a point depth (eight frequencies per coordinate).}
  \label{fig:s-scannetpp-views}
\end{figure*}

\begin{figure*}[p]
  \centering
  \includegraphics{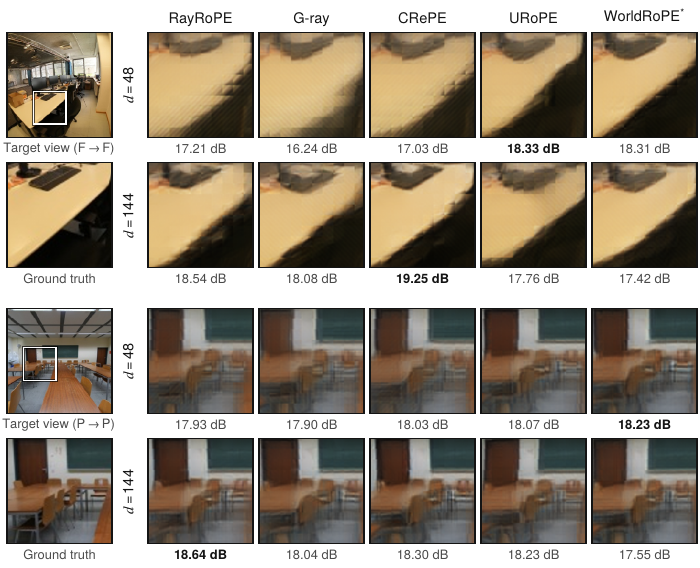}
  \caption{\textbf{Head size and method ranking on the same ScanNet++ targets.}
  Five encodings at their default bands, with 16 heads at width 768 ($d=48$) and eight heads at width 1152 ($d=144$) as in \cref{tab:s-heads} (seed 0, 20K steps). The white box marks the crop; the highest full-view PSNR of each row is in bold. For RayRoPE, G-ray on its RayRoPE host, and CRePE, the default range grows from $3\times$ to $243\times$ between the two head sizes. WorldRoPE$^\ast$ denotes the image-plane implementation. Targets are ranked by the smaller of WorldRoPE$^\ast$'s lead over RayRoPE at $d=48$ and RayRoPE's lead at $d=144$; we show the 90th percentile for fisheye and the median for pinhole.}
  \label{fig:s-scannetpp-methods}
\end{figure*}

\begin{figure*}[p]
  \centering
  \includegraphics[width=\textwidth]{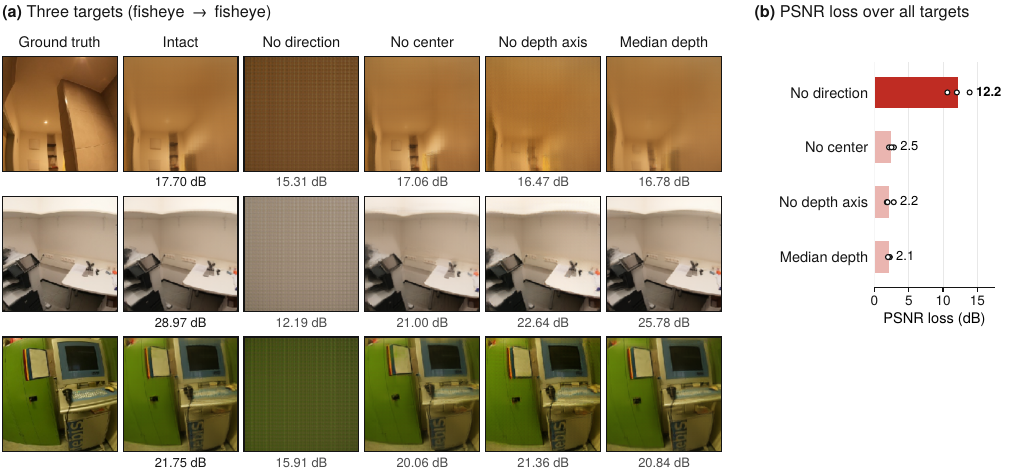}
  \caption{\textbf{Removing projected-direction rotations destroys spatial detail in trained ScanNet++ RayRoPE.}
  (a)~One $d=144$ checkpoint (seed 0, 20K steps) evaluated intact, with the rotation angles of one coordinate group set to zero, or with the patch log-depth centers replaced by their median within each view (predicted widths retained). The three scenes were drawn uniformly at random (sampling seed 0); numbers are masked full-view PSNR.
  (b)~PSNR loss relative to the intact model over all 400 fisheye-to-fisheye targets: bars average three training seeds (dots). The other three camera pairs differ from these means by at most $0.8\dB$.}
  \label{fig:s-scannetpp-probes}
\end{figure*}

\paragraph{Same-view high-frequency usage in RayRoPE.}
In 47M single-ray RayRoPE ($3^{11}$), same-view projected directions do not depend on depth, so the six frequencies above $1$ rotation/patch stay unattenuated within a view (dropping only under half-precision rounding at the top frequency, whereas cross-view amplitudes fall to $0.03$--$0.31$).
Yet zeroing those six direction frequencies on keys across 100 test scenes costs just $0.11\dB$ within a view and $0.12\dB$ across views (or $0.54\dB$ when removed from all six coordinates across all pairs, matching \cref{sec:s-analysis}).
After training with high-frequency stop-gradient, removal costs rise to $0.64\dB$ within a view and $0.27\dB$ across views.
Harmful cross-view depth gradients thus also discourage RayRoPE from using intact same-view channels, though this explains only part of the larger $-5.4\dB$ drop in footprint-free 25M WorldRoPE (\cref{fig:s-analysis}(c)), which has fewer low frequencies.

\paragraph{Training curves.}
\Cref{fig:s-train} compares training PSNR step by step under matched batch order.
At 47M, single-ray $3^{11}$ RayRoPE trails from initialization (reaching ${\sim}1.2\dB$ by 40K steps before the first run's loss jump widens the gap to $1.8\dB$; \cref{tab:s-official-seeds}), whereas rescaling to $243\times$ tracks the three-ray baseline throughout.
At 25M, default $3\times$ RayRoPE sits ${\sim}1.2\dB$ below WorldRoPE from the first few thousand steps onward, indicating a representational bottleneck rather than slow convergence.

\paragraph{Zero-shot generalization to ACID.}
To verify that our band findings hold outside the training domain, we evaluate 34 RealEstate10K-trained official-code checkpoints directly on ACID~\cite{liu2021infinite} (\num{1595} outdoor aerial scenes, two context views and three target views, no fine-tuning; \cref{tab:s-acid}).
Every main RealEstate10K trend transfers.
At 25M ($d=48$), widening RayRoPE from its three-ray $3\times$ default to four frequencies over $243\times$ adds $+0.59\dB$ ($[0.56, 0.63]$, winning on $89\%$ of scenes; $+0.61\dB$ $[0.58, 0.64]$ on seed~2), bringing RayRoPE within $0.39\dB$ of WorldRoPE while $2{,}187\times$ dips back by $0.15\dB$.
Across the fixed-width head-size sweep, RayRoPE overtakes URoPE by $+0.12\dB$ ($[0.08, 0.16]$) at $8\times96$, and rescaling $4\times192$ from $2{,}187\times$ to $243\times$ edges past WorldRoPE by $+0.08\dB$ ($[0.04, 0.13]$, $54\%$ of scenes) and leads URoPE by $+0.54\dB$.
At 47M, single-ray RayRoPE traces the same inverted-U peaking at $243\times$ ($28.42\dB$, down to $27.50\dB$ at $3\times$ and $27.18\dB$ at non-jumped $3^{11}$).
Stopping high-frequency depth gradients at $3^{11}$ recovers $90\%$ ($[88, 91]$) of the gap to $243\times$ on seed~1 ($93\%$ $[92, 94]$ against the jumped machine-A run) and $90\%$ ($[89, 92]$) on seed~2.

\section{Results on ScanNet++}
\label{sec:s-scannet}

\paragraph{Methods at $d=144$.}
\Cref{tab:s-methods} benchmarks all methods in our unified codebase at $d=144$ under the 20K schedule.
RayRoPE, RayRoPE-hosted G-ray, and CRePE lead on both datasets, followed by WorldRoPE$^\ast$, URoPE, and camera-transform encodings.
On RealEstate10K, three-ray RayRoPE leads image-plane WorldRoPE$^\ast$ by $+0.52\dB$ at 20K, compared with $+0.24\dB$ at 80K ($26.36$ vs.\ $26.13\dB$) and a tie for single-ray $243\times$ (\cref{tab:s-trained}; full WorldRoPE matches or edges past RayRoPE in \cref{tab:s-official}).
Because our unified 80K runs also receive Pl\"ucker rather than camera-frame rays at the input (\cref{sec:s-code}), training length is not the only difference between the two comparisons.
Camera distortion has little effect on relative performance: across all methods, gains over Pl\"ucker + 2D RoPE shift by at most $0.11\dB$ between fisheye and pinhole pairs, and by at most $0.25\dB$ across all four camera pairs.

\paragraph{Fractional-factorial design study.}
To produce Fig.~\mref{fig:effects}, we parameterized seven structural choices separating published methods (\cref{tab:s-factors}) and ran a resolution-IV 16-run fractional factorial (three seeds each), estimating every main effect unconfounded by two-factor interactions.
The gray shading in Fig.~\mref{fig:effects} marks single-run seed noise ($0.2\dB$ on ScanNet++, $0.13\dB$ on RealEstate10K; \cref{sec:s-impl}), which contains four of the seven choices.

The two red band bars in Fig.~\mref{fig:effects} are measured outside the factorial.
On RealEstate10K, they show official-code single-ray RayRoPE dropping by $1.02\dB$ at $d=48$ ($243\times \to 3\times$; \cref{tab:s-official}) and by $1.39\dB$ at 47M ($243\times \to 3^{11}$ on machine B; \cref{tab:s-official-seeds}).
On ScanNet++, unified-codebase single-ray RayRoPE ($d=144$, 20K) drops by $0.53\dB$ on average across three seeds when widened from $243\times$ to $3^{11}$, with nearly identical drops across the other camera pairs.
The frequency factor inside the factorial is a different test: it moves $\omega_{n-1}$ from $0.16$ to $0.64$ rotations/patch, both inside the plateau, which explains its small effect compared with leaving the plateau at $3\times$ (\cref{tab:s-otherbands}).

\paragraph{Head sizes on ScanNet++.}
\Cref{tab:s-heads} expands Tab.~\mref{tab:scannet}.
At fixed $d=96$, widening RayRoPE from $27\times$ to $243\times$ improves 20K PSNR across all four camera pairs in every seed (\cref{fig:s-scannetpp-band}), with larger early gains (${\sim}0.3$--$0.5\dB$) settling to $0.1$--$0.2\dB$ by 20K.
\Cref{fig:s-scannetpp-views} shows representative targets across quantiles of this gain distribution: the default range wins on only three of the twelve views, all at or below the 25th percentile.
Shrinking heads from $d=144$ to $48$ slightly improves every non-RayRoPE method (URoPE here assigns one depth anchor per four heads), whereas RayRoPE, RayRoPE-hosted G-ray, and CRePE drop by $1.3$--$1.4\dB$ (\cref{tab:s-heads}; qualitative comparisons in \cref{fig:s-scannetpp-methods}).
At $d=192$ ($2187\times$ default), rescaling RayRoPE to $243\times$ adds $+0.08\dB$, consistent with the gradual onset of wide-band loss in Fig.~\mref{fig:sweep}.
Finally, a compact single-ray RayRoPE rotating only 2D direction and point depth ($C=3$, giving eight frequencies per coordinate even at $d=48$) reaches $23.61\dB$ at $16\times48$ with a $243\times$ range ($+1.2\dB$ over default), supporting the guideline to drop redundant coordinates rather than truncate the band in small heads.

\begin{table}[t]
  \centering
  \caption{\textbf{High-frequency stop-gradient on ScanNet++} (unified 47M codebase, single-ray RayRoPE at $d=144$, 20K steps, 50 validation scenes).
  Seed~1 ran on the original ScanNet++ server alongside the $243\times$ runs (recovering $64\%$ $[55, 72]$ of the F$\to$F gap, $62\%$ P$\to$P, $58\%$ P$\to$F, and $65\%$ F$\to$P); seeds~0 and~2 ran as matched $3^{11}$ and stop-gradient pairs on machine~B.
  Every per-seed, per-pair 95\% scene-bootstrap CI excludes zero (F$\to$F: $[0.59, 0.77]$, $[0.34, 0.49]$, and $[0.13, 0.31]$ for seeds~0--2).}
  \label{tab:s-scannet-sg}
  \footnotesize
  \renewcommand{\arraystretch}{0.94}
  \setlength{\tabcolsep}{2.2pt}
  \begin{tabular}{@{}lcccc@{}}
\toprule
1-ray ($d{=}144$) & F$\to$F & P$\to$P & P$\to$F & F$\to$P \\
\midrule
\multicolumn{5}{@{}l}{\emph{Three-seed mean $\pm$ SD (PSNR)}} \\
$243\times$ (orig.) & 23.72$\pm$0.15 & 23.44$\pm$0.13 & 23.04$\pm$0.13 & 23.80$\pm$0.15 \\
$3^{11}$ (matched) & 23.14$\pm$0.11 & 22.91$\pm$0.08 & 22.48$\pm$0.06 & 23.22$\pm$0.12 \\
$3^{11}$, stop-grad. & 23.58$\pm$0.12 & 23.31$\pm$0.13 & 22.87$\pm$0.12 & 23.65$\pm$0.12 \\
\midrule
\multicolumn{5}{@{}l}{\emph{Paired $\Delta$PSNR (stop-gradient minus matched $3^{11}$)}} \\
Seed~0 (B) & +0.67 & +0.61 & +0.58 & +0.67 \\
Seed~1 (orig.) & +0.41 & +0.38 & +0.37 & +0.42 \\
Seed~2 (B) & +0.21 & +0.19 & +0.22 & +0.20 \\
Mean (3 seeds) & +0.43 & +0.39 & +0.39 & +0.43 \\
\bottomrule
\end{tabular}

\end{table}

\paragraph{Wide-band stop-gradient on ScanNet++.}
To test whether the high-frequency depth-gradient mechanism transfers to a second training dataset and camera model, we apply the same stop-gradient intervention (blocking depth gradients from the six frequencies above $1$ rotation/patch) to single-ray $3^{11}$ RayRoPE on ScanNet++ across three seeds (\cref{tab:s-scannet-sg}).
Every seed and camera pair improves (all 12 cells positive, each 95\% scene-bootstrap CI excluding zero), raising three-seed mean PSNR by $+0.39$ to $+0.43\dB$ across the four fisheye/pinhole pairs against the $0.49$--$0.54\dB$ wide-band drop.
On seed~1, where $243\times$, $3^{11}$, and stop-gradient ran on the same server, stop-gradient recovers $58$--$65\%$ of the gap across all four camera pairs ($64\%$ $[55, 72]$ on F$\to$F).

\paragraph{Which coordinates does the trained model use?}
In $d=144$ RayRoPE, setting projected-direction rotations to the identity at test time drops fisheye PSNR by $12.17\dB$ (three-seed mean), versus $2.48\dB$ for camera center, $2.17\dB$ for scalar depth, and $2.13\dB$ when replacing predicted log-depths with per-view medians (\cref{fig:s-scannetpp-probes}).
These tests bound reliance from above because unrotated channels stay in attention and perturb it (\cref{sec:s-analysis}); the training interventions in Sec.~\mref{sec:verification} instead isolate high-frequency depth gradients directly.

\paragraph{Dataset-dependent tuning of non-RayRoPE bands.}
Dividing $\omega_{n-1}$ by 9 across the five patch-index or camera-angle methods of \cref{tab:s-methods} (UCPE, Wei et al., GTA, PRoPE, and Pl\"ucker + 2D RoPE) improves all five on ScanNet++ by $+0.13$ to $+0.49\dB$, yet hurts all five on RealEstate10K by $-0.15$ to $-0.32\dB$.
These RealEstate10K runs use a matched bilinear loader, which raises all methods by $0.5$--$0.7\dB$ while changing inter-method gaps by under $0.2\dB$.
These five methods rotate patch indices or local ray angles rather than reprojected positions; the paper's $0.3$--$3$ rotations/patch range is for the three reprojection-based encodings on RealEstate10K (where $3\times$ is costly; \cref{tab:s-otherbands}), and we did not repeat this test for them on ScanNet++.
Fine-tuning $\omega_{n-1}$ on one dataset therefore need not transfer to another, motivating explicit reporting of standardized bands.

\paragraph{Scaling to 532M parameters.}
In a 532M model (12 layers, width 3072, $d=192$, 80K steps, one seed), RayRoPE still outperforms URoPE on ScanNet++ ($25.87$ vs.\ $25.57\dB$), though by a narrower margin than at 47M; the band was not varied at this scale.

\section{Additional Analysis}
\label{sec:s-analysis}

\subsection{Gradient of an Averaged Rotation}

Here we derive the window-smoothing and fixed-kernel gradient bounds referenced in Sec.~\mref{sec:analysis}.
Consider a scalar coordinate $x = \mu + \xi$, where $\mu$ depends on predicted depth and $\xi$ is a zero-mean offset with density $p$ and characteristic function $\hat p(\omega) = \mathbb{E}[e^{i\omega\xi}]$.
Holding $p$ fixed, the expected rotation and the magnitude of its $\mu$-derivative are
\begin{equation}
  \Phi_\omega = \mathbb{E}[e^{i\omega x}] = e^{i\omega\mu}\,\hat p(\omega),
  \quad
  \Bigl|\frac{\partial \Phi_\omega}{\partial \mu}\Bigr| = \omega\,\bigl|\hat p(\omega)\bigr|.
  \label{eq:s-grad}
\end{equation}
Thus forward outputs scale with $|\hat p(\omega)|$, whereas gradients with respect to $\mu$ scale with $\omega\,|\hat p(\omega)|$.

\paragraph{Uniform interval.}
For the uniform density on $[-\Delta, \Delta]$ with fixed $\Delta>0$ (Eq.~(\mref{eq:box})), $\hat p(\omega) = \sin(\omega\Delta)/(\omega\Delta)$, leaving an undamped $\mu$-gradient magnitude $|\sin\omega\Delta|/\Delta = O(1)$.
Its derivative with respect to the half-width, $\partial_\Delta \Phi_\omega = e^{i\omega\mu}\,\bigl(\cos(\omega\Delta) - \hat p(\omega)\bigr)/\Delta$, does not decay either: as $\omega \to \infty$ at fixed $\Delta>0$, the forward rotation vanishes while both derivatives keep nondecaying oscillatory envelopes of order $1/\Delta$.

At the pair level, fix query token $i$'s camera frame and write $\Phi_i(\omega), \Phi_j(\omega) \in \mathbb{C}$ for query $i$'s and key/value $j$'s averaged coefficients in that frame.
RayRoPE multiplies queries, keys, and values by $\overline{\Phi}$ and attention outputs by $\Phi$, so each complex channel pair $q_{i,\omega}, k_{j,\omega}, v_{j,\omega} \in \mathbb{C}$ enters the attention logit as $\operatorname{Re}\bigl[\Gamma_{ij}(\omega)\,\overline{q_{i,\omega}}k_{j,\omega}\bigr]$ (up to logit scale) and the transported output as $\sum_j A_{ij}\Gamma_{ij}(\omega)v_{j,\omega}$ via the pair coefficient $\Gamma_{ij}(\omega) := \Phi_i(\omega)\overline{\Phi_j(\omega)}$.
Consequently, the $O(1)$ center-derivative envelope of $\Phi_j$ passes directly to $\Gamma_{ij}$ whenever the paired side $\Phi_i$ is unattenuated (for example, on projected image-plane coordinates where query $i$'s ray in its own camera has zero depth uncertainty, or when one token's interval is narrow).
When both sides have fixed nonzero widths, the fixed-window pair derivative scales as $O(1/\omega)$.
We state this for RayRoPE's uniform window and do not treat other discontinuous windows.

\begin{table*}[t]
  \centering
  \caption{\textbf{Exploratory wide-band modifications in the 47M analysis model} (unified codebase, mean validation PSNR over 10K--20K steps; recovery defined in \cref{sec:s-impl}).
  Frequencies are indexed $0$ to $11$; under $3^{11}$, indices $6$--$11$ exceed $1$ rotation per patch.
  Full 80K official-code counterparts appear in Fig.~\mref{fig:mechanism}(c) and \cref{tab:s-official-mods}.}
  \label{tab:s-modifications}
  \footnotesize
  \renewcommand{\arraystretch}{0.95}
  \setlength{\tabcolsep}{6pt}
  \begin{tabular}{@{}lrrr@{}}
\toprule
Model & PSNR & vs.\ $243\times$ & Recovery [95\% CI] \\
\midrule
\multicolumn{4}{@{}l}{\emph{Frequency range}} \\
\quad range $3^{11}$ (default rule) & 23.36 & $-$0.88 & 0\% (def.) \\
\quad range $3^{11}$, second run & 23.39 & $-$0.85 & 3\% [$-$1, 8] \\
\quad range $3^{9}$ & 23.97 & $-$0.27 & 69\% [65, 73] \\
\quad range $3^{7}$ & 24.13 & $-$0.11 & 87\% [83, 91] \\
\quad range $243\times$ & 24.24 & -- & 100\% (def.) \\
\quad range $3^{3}$ & 23.82 & $-$0.42 & 52\% [48, 57] \\
\quad range $3^{1}$ & 22.75 & $-$1.49 & $-$70\% [$-$82, $-$59] \\
\multicolumn{4}{@{}l}{\emph{Stop the depth gradient of some frequencies (range $3^{11}$)}} \\
\quad frequencies 6--11 & 24.14 & $-$0.10 & 88\% [84, 92] \\
\quad frequencies 9--11 & 24.05 & $-$0.19 & 79\% [75, 83] \\
\quad frequencies 10--11 & 23.81 & $-$0.43 & 51\% [47, 54] \\
\quad frequencies 6--8 & 23.24 & $-$1.00 & $-$14\% [$-$18, $-$10] \\
\quad frequencies 0--5 & 20.89 & $-$3.35 & $-$281\% [$-$300, $-$264] \\
\quad frequencies 6--11, through the interval center only & 23.81 & $-$0.43 & 51\% [48, 54] \\
\quad frequencies 6--11, through the interval width only & 23.36 & $-$0.88 & 0\% [$-$4, 3] \\
\multicolumn{4}{@{}l}{\emph{Controls (range $3^{11}$)}} \\
\quad no gradient clipping & 23.31 & $-$0.93 & $-$6\% [$-$10, $-$1] \\
\quad stop-gradient on frequencies 6--11, no gradient clipping & 24.12 & $-$0.12 & 87\% [83, 91] \\
\quad all depth gradients $\times0.5$ & 23.80 & $-$0.44 & 50\% [45, 54] \\
\quad depth gradient of frequencies 6--11 $\times0.5$ & 23.70 & $-$0.54 & 38\% [35, 42] \\
\quad remove frequencies 6--11 from the encoding & 23.97 & $-$0.27 & 69\% [65, 72] \\
\quad range $243\times$, stop-gradient on frequencies 6--11 & 23.60 & $-$0.65 & 27\% [21, 33] \\
\multicolumn{4}{@{}l}{\emph{Stop-gradient above a fixed frequency}} \\
\quad range $3^{7}$, frequencies $\geq 3^{6}\omega_0$ (10--11) & 24.19 & $-$0.05 & 95\% [90, 99] \\
\quad range $3^{9}$, frequencies $\geq 3^{6}\omega_0$ (8--11) & 24.06 & $-$0.18 & 80\% [76, 83] \\
\quad range $3^{9}$, frequencies $\geq 3^{9}\omega_0$ (11) & 24.12 & $-$0.12 & 86\% [81, 90] \\
\multicolumn{4}{@{}l}{\emph{Ranges that extend $243\times$ upward}} \\
\quad 8 frequencies over $243\times$ and 4 above (range $3^{7.9}$) & 24.03 & $-$0.21 & 76\% [72, 80] \\
\quad \quad with stop-gradient on the 4 upper frequencies & 24.17 & $-$0.07 & 92\% [88, 96] \\
\quad 10 frequencies over $243\times$ and 2 above (range $3^{6.1}$) & 24.26 & +0.02 & 102\% [98, 107] \\
\quad \quad with stop-gradient on the 2 upper frequencies & 24.29 & +0.05 & 106\% [102, 111] \\
\bottomrule
\end{tabular}

\end{table*}

\paragraph{Smooth learned windows.}
If $p$ has $m$ integrable derivatives, $|\hat p(\omega)| \le \|p^{(m)}\|_1\,\omega^{-m}$, so $\omega\,|\hat p(\omega)| \to 0$ whenever $m \ge 2$ (for a Gaussian of width $\sigma$, it equals $\omega\,e^{-\omega^2\sigma^2/2}$; cf.\ WorldRoPE's forward displacement bound~\cite{worldrope2026}, Eq.~12).
However, if $\sigma$ is predicted within a scale family $\hat p_\sigma(\omega) = \hat p_1(\sigma\omega)$, shrinking $\sigma \to 0$ reactivates arbitrarily high frequencies in both forward outputs and backward gradients.

\paragraph{Fixed-width query-plane smoothing kernel (and distinction from WorldRoPE).}
Adding an independent, non-learned offset $\eta \sim g$ directly in the projected coordinate ($x = \mu + \xi + \eta$) multiplies the characteristic function by $\hat g(\omega)$: $\Phi_\omega = e^{i\omega\mu}\,\hat p(\omega)\,\hat g(\omega)$.
Because $|\hat p(\omega)| \le 1$ for any distribution $p$,
\begin{equation}
  \Bigl|\frac{\partial \Phi_\omega}{\partial \mu}\Bigr| \le \omega\,\bigl|\hat g(\omega)\bigr|
  \label{eq:s-floor}
\end{equation}
uniformly over all predicted windows $p$.
This bounds the derivative with respect to center $\mu$ (not directly with respect to window width or the full projection path to the depth predictor).
In our variance-floor interventions below, $g$ is a fixed Gaussian matching one uniform patch ($\sigma_0^2 = 1/12$ patch$^2$), whose gradient bound peaks near $0.5$ rotations/patch and falls below $1\%$ of its peak beyond $2$ rotations/patch.
Note that $g$ itself must be smooth: convolving with a fixed uniform box still has an $O(1/\omega)$ envelope and fails to damp gradients.
Note also the difference from WorldRoPE~\cite{worldrope2026}: WorldRoPE fixes the \emph{source}-patch covariance ($\mathrm{diag}(1/12, 1/12)$ patch$^2$) and projects it through 3D into the query view, so its cross-view query-plane variance still varies with depth and pose, reducing to a constant $\sigma_0^2 = 1/12$ kernel only within the \emph{same} view (\cref{sec:s-re10k}).

\subsection{Modifications of the Wide Band}

\Cref{tab:s-modifications} reports all exploratory modifications in the 47M analysis model.
Throughout, \emph{stopping} the depth gradient of a set of frequencies detaches their projected coordinates from the computation graph across all rotated coordinates, layers, and token pairs: the forward pass is untouched, and no gradient flows from those coefficients to predicted depth center or width (unless one is specified).
\emph{Scaling} multiplies that gradient by a constant instead.

\paragraph{Which frequencies and paths cause harm.}
Stopping depth gradients on only the top three frequencies ($9$--$11$) recovers $79\%$ of the gap, whereas stopping indices $6$--$8$ alone helps not at all, so the damage concentrates at the top end.
Conversely, stopping indices $6$--$11$ in the $243\times$ model (where they lie \emph{below} $1$ rotation/patch) hurts PSNR by $0.65\dB$, confirming that physical frequency rather than channel index drives the effect.
Along the interval parameterization, blocking gradients through the center $\mu$ alone recovers $51\%$, blocking half-width $\Delta$ alone recovers nothing, and blocking both is needed for full recovery ($88\%$).

\paragraph{Coordinate-group screens.}
Widening one coordinate group at a time to $3^{11}$ (keeping others at $243\times$) shows that projected ray direction causes the largest drop, scalar depth a smaller drop, and camera center the smallest, with individual drops summing to more than joint widening.
Replacing the box interval with a learned-width Gaussian mitigates the wide-depth drop without reducing the wide-direction drop.
Conversely, narrowing projected direction alone to $3\times$ reproduces the entire narrow-band loss, while narrowing camera center costs nothing.
Full 80K official-code counterparts appear in \cref{tab:s-official-groups}.

\paragraph{Gradient clipping, scaling, and upward band extension.}
Disabling gradient clipping alters neither the $3^{11}$ baseline nor stop-gradient recovery.
Halving all depth gradients recovers $50\%$ of the gap, while halving only indices $6$--$11$ recovers $38\%$; because AdamW is nearly invariant to uniform rescaling of the depth predictor's own gradients, halving all depth gradients acts mainly on the branch flowing into shared features.
Finally, appending two extra frequencies above a $243\times$ band (${\sim}3$ rotations/patch) causes no loss with or without stop-gradient, so we view stop-gradient strictly as a diagnostic tool rather than a modeling proposal.

\paragraph{High-frequency forward usage vs.\ gradient share.}
In Fig.~\mref{fig:mechanism}(b), zeroing any single key frequency above $1$ rotation/patch in $3^{11}$ RayRoPE raises MSE by at most $6\%$ across three official runs (and $\le 2.6\%$ in the analysis model, aside from the boundary frequency), whereas zeroing any single lower frequency raises MSE by $36\%$ to nearly $1000\%$.
(Setting rotations to the identity instead of zeroing keys produces larger drops because unrotated channels actively perturb attention.)
Despite this minimal forward role, frequencies above $1$ rotation/patch carry $19$--$49\%$ of summed per-frequency depth-gradient norms in official $3^{11}$ RayRoPE runs (and $28$--$70\%$ in the analysis model), matching \cref{eq:s-grad}.
By contrast, in $3^{11}$ WorldRoPE, frequencies above $1$ rotation/patch at 25M and above $3$ rotations/patch at 47M each alter MSE by under $0.3\%$ and together carry under $1\%$ of depth-gradient norms.

\begin{table*}[t]
  \centering
  \caption{\textbf{Wide-band ($3^{11}$) modifications in the official codebase} (47M single-ray RayRoPE, RealEstate10K test PSNR at 80K steps; machines A--C).
  \emph{Recovery}: fraction of the gap to $243\times$ ($26.61\dB$ seed~1, $26.66\dB$ seed~2, $26.64\dB$ seed~3) closed relative to the matched-seed unmodified $3^{11}$ run on the same machine (seed~2 on C paired with seed~2 on B).
  $^\ast$Trained to 34K on C and resumed to 80K on B.
  $^\ddagger$Gain over matched-seed single-precision geometry alone ($25.70$, $25.67$, $25.74$, $25.51\dB$): $+0.70$, $+0.87$, $+0.75$, $+0.87\dB$, closing $77\%$, $92\%$, $81\%$, and $76\%$ of the remaining gap.
  $^\S$Gain over same-machine single-precision geometry ($25.67\dB$ on C; $25.79\dB$ on A): $+0.61$, $+0.26$, $+0.65$, $-0.25$, $+0.07\dB$ ($65\%$, $28\%$, $78\%$, $-31\%$, and $9\%$ of the remaining gap).}
  \label{tab:s-official-mods}
  \footnotesize
  \renewcommand{\arraystretch}{0.94}
  \setlength{\tabcolsep}{6pt}
  \begin{tabular}{@{}llrrrr@{}}
\toprule
Modification of the $3^{11}$ band & Machine & PSNR & $\Delta$ to $3^{11}$ [95\% CI] & $\Delta$ to $243\times$ [95\% CI] & Recovery [95\% CI] \\
\midrule
\multicolumn{6}{@{}l}{\emph{Unmodified $3^{11}$ baselines (default $3^k$ rule, seed 1)}} \\
unmodified $3^{11}$, seed 1 & A & 24.54 & -- & $-$2.07 [$-$2.10, $-$2.04] & -- \\
\quad seed 1 rerun on B & B & 25.22 & -- & $-$1.39 [$-$1.41, $-$1.37] & -- \\
\quad seed 1 rerun on C & C & 25.41 & -- & $-$1.20 [$-$1.22, $-$1.17] & -- \\
\midrule
\multicolumn{6}{@{}l}{\emph{Mixed-precision interventions (seed 1)}} \\
stop depth gradient of frequencies 6--11 & A & 26.47 & +1.93 [1.90, 1.96] & $-$0.14 [$-$0.16, $-$0.13] & 93\% [92, 94] \\
\quad rerun on C (vs.\ $25.41\dB$) & C & 26.47 & +1.05 [1.03, 1.07] & $-$0.14 [$-$0.16, $-$0.13] & 88\% [87, 89] \\
\quad values and outputs only$^\ast$ & C, B & 26.15 & +0.74 [0.72, 0.75] & $-$0.46 [$-$0.47, $-$0.44] & 62\% [61, 63] \\
\quad values and outputs only (on C) & C & 25.86 & +0.45 [0.43, 0.46] & $-$0.75 [$-$0.77, $-$0.73] & 37\% [36, 38] \\
\quad queries and keys only & C & 25.05 & $-$0.37 [$-$0.38, $-$0.35] & $-$1.56 [$-$1.59, $-$1.53] & $-$31\% [$-$32, $-$29] \\
stop depth gradient of frequencies 9--11 & B & 26.33 & +1.11 [1.09, 1.13] & $-$0.28 [$-$0.29, $-$0.27] & 80\% [79, 81] \\
stop depth gradient of frequencies 0--5 & B & 21.74 & $-$3.48 [$-$3.53, $-$3.43] & $-$4.87 [$-$4.93, $-$4.81] & $-$250\% [$-$255, $-$245] \\
remove frequencies 6--11 from encoding & B & 26.50 & +1.28 [1.26, 1.30] & $-$0.11 [$-$0.12, $-$0.09] & 92\% [91, 93] \\
layer $l$ uses the $2l$ lowest frequencies & B & 26.31 & +1.10 [1.08, 1.11] & $-$0.30 [$-$0.31, $-$0.28] & 79\% [78, 80] \\
all depth gradients $\times 0.5$ & B & 25.83 & +0.61 [0.60, 0.63] & $-$0.78 [$-$0.80, $-$0.76] & 44\% [43, 45] \\
Gaussian window $+$ fixed one-patch floor & B & 26.32 & +1.10 [1.08, 1.12] & $-$0.29 [$-$0.31, $-$0.28] & 79\% [78, 80] \\
interval window $+$ fixed one-patch floor & B & 26.22 & +1.00 [0.98, 1.02] & $-$0.39 [$-$0.41, $-$0.38] & 72\% [71, 73] \\
Gaussian window, learned width & B & 25.59 & +0.37 [0.36, 0.39] & $-$1.02 [$-$1.04, $-$1.00] & 27\% [26, 28] \\
training entirely in single precision & B & 25.64 & +0.42 [0.40, 0.43] & $-$0.97 [$-$0.99, $-$0.95] & 30\% [29, 31] \\
\midrule
\multicolumn{6}{@{}l}{\emph{Single-precision geometry controls and combined interventions}} \\
geometry in single precision, seed 1 & B & 25.70 & +0.48 [0.47, 0.50] & $-$0.91 [$-$0.93, $-$0.89] & 35\% [34, 36] \\
\quad seed 1 on A (vs.\ $24.54\dB$) & A & 25.79 & +1.25 [1.23, 1.27] & $-$0.82 [$-$0.84, $-$0.81] & 60\% [59, 61] \\
\quad seed 1 on C (vs.\ $25.41\dB$) & C & 25.67 & +0.25 [0.24, 0.27] & $-$0.94 [$-$0.96, $-$0.92] & 21\% [20, 22] \\
\quad seed 2 (vs.\ $25.31$ / $26.66\dB$) & B & 25.74 & +0.42 [0.41, 0.44] & $-$0.92 [$-$0.94, $-$0.90] & 31\% [30, 32] \\
\quad seed 3 (vs.\ $23.56$ / $26.64\dB$) & C & 25.51 & +1.95 [1.91, 1.98] & $-$1.13 [$-$1.16, $-$1.11] & 63\% [63, 64] \\
\quad $+$ stop depth grad.\ (6--11), seed 1$^\ddagger$ & B & 26.40 & +1.19 [1.16, 1.21] & $-$0.21 [$-$0.22, $-$0.19] & 85\% [84, 86] \\
\quad $+$ stop depth grad.\ (6--11), seed 1 on C$^\ddagger$ & C & 26.54 & +1.12 [1.10, 1.14] & $-$0.07 [$-$0.09, $-$0.06] & 94\% [93, 95] \\
\quad $+$ stop depth grad.\ (6--11), seed 2$^\ddagger$ & C & 26.49 & +1.17 [1.15, 1.20] & $-$0.17 [$-$0.18, $-$0.16] & 87\% [86, 88] \\
\quad $+$ stop depth grad.\ (6--11), seed 3$^\ddagger$ & C & 26.37 & +2.81 [2.77, 2.85] & $-$0.27 [$-$0.28, $-$0.25] & 91\% [91, 92] \\
\quad $+$ stop values and outputs only$^\S$ & C & 26.28 & +0.87 [0.85, 0.88] & $-$0.33 [$-$0.34, $-$0.32] & 72\% [71, 73] \\
\quad $+$ stop queries and keys only$^\S$ & C & 25.93 & +0.51 [0.50, 0.53] & $-$0.68 [$-$0.70, $-$0.67] & 43\% [42, 44] \\
\quad $+$ stop frequencies 9--11 only$^\S$ & A & 26.43 & +1.89 [1.87, 1.92] & $-$0.18 [$-$0.19, $-$0.16] & 91\% [91, 92] \\
\quad $+$ stop frequencies 6--8 only$^\S$ & A & 25.53 & +1.00 [0.97, 1.02] & $-$1.08 [$-$1.10, $-$1.06] & 48\% [47, 49] \\
\quad $+$ Gaussian window, learned width$^\S$ & A & 25.86 & +1.32 [1.30, 1.35] & $-$0.75 [$-$0.77, $-$0.73] & 64\% [63, 64] \\
\midrule
\multicolumn{6}{@{}l}{\emph{Reference $243\times$ runs (seed 1)}} \\
range $243\times$ (mixed precision) & A & 26.61 & -- & -- & -- \\
\quad rerun on C & C & 26.57 & -- & $-$0.04 [$-$0.05, $-$0.03] & -- \\
\quad geometry in single precision & C & 26.62 & -- & +0.01 [0.00, 0.02] & -- \\
\bottomrule
\end{tabular}

\end{table*}

\paragraph{Fixed variance-floor intervention in the analysis model.}
In the analysis model (second seed), replacing RayRoPE's box interval with an equal-variance learned-width Gaussian recovers only $8\%$ of the $3^{11}$ gap ($4\%$ on the first seed).
This is not due to width collapse: the learned RMS extent in projected direction stays near $0.09$ in the last layer and grows from $0.19$ to $0.29$ in the first (though this average does not rule out a narrow-window tail).
Adding our detached one-patch Gaussian variance floor ($\sigma_0^2 = 1/12$ patch$^2$, applied on both queries and keys) raises recovery to $42\%$ ($+0.36\dB$) with a Gaussian window and $67\%$ ($+0.59\dB$) with RayRoPE's original interval.
Consistent with the convolution bound of \cref{eq:s-floor}, the fixed smooth floor rather than a Gaussian base window drives the recovery.

\paragraph{Full 80K modifications in the official codebase.}
\Cref{tab:s-official-mods} reports the full 80K interventions on 47M single-ray RayRoPE.
Blocking depth gradients on the six highest frequencies recovers $88$--$93\%$ of the gap across seeds and machines (\cref{tab:s-official-seeds}), matching complete removal of those frequencies ($92\%$).
Stopping only the top three already recovers ${\sim}80\%$, whereas stopping the middle three ($6$--$8$) lowers single-precision-geometry PSNR by $0.25\dB$ (consistent with the analysis model).
Stopping the six \emph{lowest} frequencies stalls training ($21.74\dB$), and halving all depth gradients recovers less than half the gap.

\begin{table*}[t]
  \centering
  \caption{\textbf{Varying one coordinate group's band at a time in the official codebase} (47M single-ray RayRoPE, 80K steps, one seed).
  \emph{Share of the loss}: fraction of the all-group degradation caused by changing that group alone (relative to the machine-B $3^{11}$ run at $25.22\dB$ in the top block, and $3\times$ in the bottom block).}
  \label{tab:s-official-groups}
  \footnotesize
  \setlength{\tabcolsep}{6pt}
  \begin{tabular}{@{}llllrrr@{}}
\toprule
Direction & Camera center & Depth & Trained on & PSNR & $\Delta$ to $243\times$ [95\% CI] & Share of the loss [95\% CI] \\
\midrule
$243\times$ & $243\times$ & $243\times$ & A & 26.61 & -- & -- \\
$3^{11}$ & $3^{11}$ & $3^{11}$ & B & 25.22 & $-$1.39 [$-$1.41, $-$1.37] & -- \\
$3^{11}$ & $3^{11}$ & $3^{11}$ & A & 24.54 & $-$2.07 [$-$2.10, $-$2.04] & -- \\
$3^{11}$ & $243\times$ & $243\times$ & B & 25.44 & $-$1.17 [$-$1.19, $-$1.14] & 84\% [83, 85] \\
$243\times$ & $3^{11}$ & $243\times$ & B & 26.51 & $-$0.10 [$-$0.12, $-$0.09] & 7\% [7, 8] \\
$243\times$ & $243\times$ & $3^{11}$ & B & 26.56 & $-$0.05 [$-$0.06, $-$0.04] & 4\% [3, 4] \\
\midrule
$3^{1}$ & $3^{1}$ & $3^{1}$ & B & 25.38 & $-$1.23 [$-$1.26, $-$1.21] & -- \\
$3^{1}$ & $243\times$ & $243\times$ & B & 25.47 & $-$1.14 [$-$1.17, $-$1.12] & 93\% [92, 94] \\
\bottomrule
\end{tabular}

\end{table*}

Splitting stop-gradient by pathway shows that values and outputs carry most of the damage.
In mixed precision, blocking queries and keys alone gives no recovery, while blocking values and outputs recovers roughly $40$--$60\%$ across two runs (which separate around 30K steps from run-to-run variance).
Under single-precision geometry, the two routes are nearly additive: stopping values/outputs closes roughly two thirds of the remaining gap, stopping queries/keys closes roughly a quarter, and stopping both closes $92\%$ (\cref{tab:s-official-mods}).
A trained-model gradient probe matches this split, with high-frequency gradients dividing $23\%$ through values/outputs versus $3\%$ through queries/keys.

Among non-stop-gradient controls, adding the fixed one-patch Gaussian floor or a coarse-to-fine layer schedule ($2l$ lowest frequencies in layer $l$) closes $72$--$79\%$ of the gap; a learned-width Gaussian alone closes only $27\%$ in mixed precision and adds little under single-precision geometry.
Single-precision geometry alone closes $21$--$63\%$ depending on the baseline (detailed below).
Despite differing in step budget, input rays, and depth coordinate, both codebases produce the same ordering across the main intervention groups: stopping or removing high frequencies near $90\%$, fixed floors in between, and a learned width alone at the bottom.

\paragraph{Coordinate groups in the official codebase.}
\Cref{tab:s-official-groups} modifies one coordinate group at a time in 47M single-ray RayRoPE.
Narrowing only projected direction to $3\times$ reproduces $93\%$ of the narrow-band loss.
Widening only projected direction to $3^{11}$ accounts for $84\%$ of the wide-band loss, versus $7\%$ for camera center and $4\%$ for depth; the three single-group drops sum to $95\%$, acting nearly additively.
Although the two small groups reverse order relative to the 20K analysis screen, both settings confirm that projected direction dominates both regimes.

\begin{table*}[t]
  \centering
  \caption{\textbf{Wide-band degradation and stop-gradient recovery across backbone shapes} in the unified codebase (mean validation PSNR over 10K--20K steps; stop-gradient blocks all frequencies above the $243\times$ ceiling).}
  \label{tab:s-shapes}
  \footnotesize
  \renewcommand{\arraystretch}{0.94}
  \begin{tabular}{@{}lcrrrc@{}}
\toprule
Model & Freq. & $3^{11}$ & $243\times$ & $3^{11}$, stop-gradient & Recovery [95\% CI] \\
\midrule
12 layers & 12 & 23.40 & 24.77 & 24.60 & 88\% [85, 90] \\
4 heads of 288 channels & 24 & 23.28 & 24.03 & 23.91 & 84\% [79, 89] \\
width 576, 4 heads & 12 & 23.31 & 23.91 & 23.74 & 72\% [66, 77] \\
16 heads of 72 channels & 6 & 23.32 & 24.29 & 23.93 & 63\% [60, 66] \\
\bottomrule
\end{tabular}

\end{table*}

\begin{table}[t]
  \centering
  \caption{\textbf{Finer range sweep across 8- and 16-pixel patches} in the analysis model (one seed, validation PSNR over 10K--20K steps; \emph{Highest}: direction $\omega_{n-1}$ in rotations/patch; $\Delta$PSNR relative to $243\times$ at the same patch size).}
  \label{tab:s-patch}
  \footnotesize
  \setlength{\tabcolsep}{4pt}
  \begin{tabular}{@{}lrrrr@{}}
    \toprule
    & \multicolumn{2}{c}{8-pixel patches} & \multicolumn{2}{c}{16-pixel patches} \\
    Range & Highest & $\Delta$PSNR & Highest & $\Delta$PSNR \\
    \midrule
    $3^{1}$ & 0.01 & $-1.49$ & 0.02 & $-1.28$ \\
    $3^{3}$ & 0.11 & $-0.42$ & 0.21 & $-0.36$ \\
    $3^{4}$ & 0.32 & $-0.06$ & 0.63 & $-0.20$ \\
    $3^{4.5}$ & 0.55 & $-0.03$ & 1.10 & $-0.02$ \\
    $3^{5} = 243\times$ & 0.95 & 24.24\dB & 1.90 & 22.96\dB \\
    $3^{5.25}$ & -- & -- & 2.5 & $+0.11$ \\
    $3^{5.5}$ & 1.64 & $+0.08$ & 3.3 & $+0.20$ \\
    $3^{5.75}$ & -- & -- & 4.3 & $+0.18$ \\
    $3^{6}$ & 2.85 & $+0.08$ & 5.7 & $+0.06$ \\
    $3^{6.5}$ & -- & -- & 9.9 & $+0.03$ \\
    $3^{7}$ & 8.5 & $-0.11$ & 17 & $-0.02$ \\
    $3^{9}$ & 77 & $-0.27$ & 154 & $-0.21$ \\
    $3^{11}$ & 692 & $-0.88$ & 1{,}384 & $-0.70$ \\
    \bottomrule
  \end{tabular}
\end{table}

\begin{figure*}[t]
  \centering
  \includegraphics[width=0.95\textwidth]{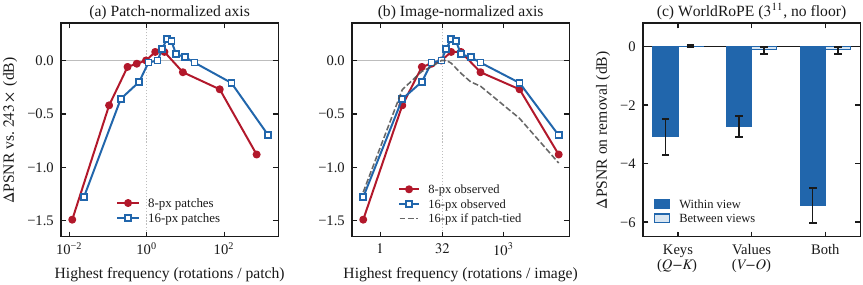}
  \caption{\textbf{Image-scale alignment and same-view channel usage.}
  (a,\,b)~Analysis-model $\Delta$PSNR vs.\ $243\times$ for 8- and 16-pixel patches (\cref{tab:s-patch}) plotted against rotations per patch~(a) and rotations per image~(b), with the $2\times$ shift predicted by a patch-tied rule (dashed).
  (c)~Test PSNR drop on 32 scenes when removing the four image-plane frequencies $>1$ rotation/patch from 25M $3^{11}$ WorldRoPE trained with its source-patch footprint omitted (``no floor'' in panel title), evaluated on keys, values, or both.}
  \label{fig:s-analysis}
\end{figure*}

\paragraph{Finer sweep and patch size in the analysis model.}
\Cref{tab:s-patch} and \cref{fig:s-analysis}(a,\,b) sample the plateau densely around $243\times$.
With 8-pixel patches, ranges from $3^4$ to $3^6$ stay within $0.15\dB$ of one another (vs.\ up to $0.06\dB$ rerun noise).
Doubling patch size to 16 pixels aligns more closely with image coordinates than with a $2\times$ downward shift (dashed in \cref{fig:s-analysis}(b)): wide-band drops shrink rather than grow, narrow-band losses appear earlier, and the peak remains just above $243\times$.
The finite grid and single-seed runs still leave the exact scaling law open.

\paragraph{Other backbone shapes and full 80K unified runs.}
Across four altered depth, width, and head-count layouts (\cref{tab:s-shapes}), $3^{11}$ trails $243\times$ by $0.6$--$1.4\dB$ and stop-gradient recovers $63$--$88\%$.
\begin{table*}[t]
  \centering
  \caption{\textbf{Full 80K RealEstate10K test results in our unified 47M codebase} (seed~2), including WorldRoPE$^\ast$ and URoPE with upward-shifted bands ($^\ast$image-plane coordinates only, \cref{sec:s-code}).}
  \label{tab:s-trained}
  \footnotesize
  \renewcommand{\arraystretch}{0.94}
  \begin{tabular}{@{}lrlr@{}}
\toprule
Model & PSNR & Difference & [95\% CI] \\
\midrule
RayRoPE, range $3^{11}$ & 25.38 &  &  \\
RayRoPE, range $243\times$ & 26.29 & vs.\ $3^{11}$: +0.92 & [0.90, 0.93] \\
RayRoPE, range $3^{11}$, stop-gradient & 26.13 & vs.\ $3^{11}$: +0.75 & [0.73, 0.77] \\
RayRoPE, 10 frequencies over $243\times$ and 2 above & 26.35 & vs.\ $243\times$: +0.06 & [0.04, 0.07] \\
\quad with stop-gradient on the 2 upper frequencies & 26.37 & vs.\ $243\times$: +0.08 & [0.07, 0.09] \\
WorldRoPE$^\ast$, default frequencies & 26.29 & vs.\ RayRoPE $3^{11}$: +0.92 & [0.90, 0.94] \\
WorldRoPE$^\ast$, 0.01--0.5 rotations per patch & 26.33 & vs.\ default: +0.04 & [0.03, 0.05] \\
URoPE, default frequencies & 25.81 & vs.\ RayRoPE $3^{11}$: +0.43 & [0.41, 0.45] \\
URoPE, 0.01--0.5 rotations per patch & 26.04 & vs.\ default: +0.23 & [0.22, 0.24] \\
\midrule
\multicolumn{4}{@{}l}{Recovery of the stop-gradient on the test set: 82\% [81, 83]} \\
\bottomrule
\end{tabular}

\end{table*}
Full 80K unified-codebase runs (\cref{tab:s-trained}) confirm the 20K screens ($82\%$ stop-gradient recovery on seed~2, $89\%$ on seed~1).
Shifting image-plane bands upward to $0.01$--$0.5$ rotations/patch improves URoPE by $+0.23\dB$ here while leaving WorldRoPE$^\ast$ nearly flat ($+0.04\dB$); in the official codebase (\cref{tab:s-otherbands}), however, raising either method to $1$ rotation/patch stays within default seed variation.
Because the two codebases also differ in input rays, depth-anchor count (two vs.\ four at eight heads), and $\omega_0$, whether $0.16$ rotations/patch already lies on URoPE's plateau remains open, and we did not sample between $0.012$ and $0.16$ rotations/patch to locate the exact lower edge.

\paragraph{Per-frequency probe protocol.}
In Fig.~\mref{fig:mechanism}(b) (\emph{top}), we zero key-side rotation coefficients for one frequency across all layers (all coordinates in RayRoPE; image-plane coordinates in WorldRoPE) and record the relative MSE increase on 30 test scenes for RayRoPE, 100 for WorldRoPE, and 24 validation scenes for the analysis model.
In Fig.~\mref{fig:mechanism}(b) (\emph{bottom}), we run one training-loss backward pass (24 test scenes for RayRoPE, 32 for WorldRoPE), isolate the gradient through a single frequency's rotation coefficients, and propagate it to that layer's depth-predictor parameters.
Because parameter gradients are linear in upstream cotangents, per-frequency vectors sum to the total gradient; each frequency's share is its scene-averaged norm divided by the sum of scene-averaged norms across all frequencies, accumulated tail-wise in Fig.~\mref{fig:mechanism}(b) (norm shares, not fractions of the summed-vector norm, since contributions can cancel).

\paragraph{Learned-width Gaussians: RayRoPE vs.\ WorldRoPE.}
In \cref{sec:s-re10k} and the paper's Limitations, we noted that a learned-width Gaussian tolerates $3^{11}$ much better in WorldRoPE without its footprint ($26.95\dB$; \cref{tab:s-otherbands}) than in RayRoPE ($25.59\dB$, $27\%$ recovery; \cref{tab:s-official-mods}).
To inspect this difference, we measure learned cross-view window widths, high-frequency removal cost, and per-frequency gradient shares on both trained 47M checkpoints.
Cross-view window widths do not explain the gap: across the six layers, median projected cross-view $\sigma$ spans $0.02$--$1.06$ patches (for tokens projecting inside the query frame) in Gaussian RayRoPE, comparable to or wider than WorldRoPE's $0.06$--$0.38$ patches (context-to-context) and $0.03$--$0.15$ patches (context-to-target).
Forward reliance and gradient concentration, however, differ sharply.
Zeroing frequencies above $1$ rotation/patch at test time reduces PSNR by only $-0.41\dB$ in Gaussian RayRoPE ($30$ scenes), versus $-7.0\dB$ in WorldRoPE ($100$ scenes, driven by same-view channels).
Conversely, frequencies above $3$ rotations/patch carry $35\%$ of depth-gradient norm in Gaussian RayRoPE ($44\%$ above $1$ rotation/patch and $16\%$ at $692$ rotations/patch alone, $24$ scenes), versus only $6.0\%$ of mixture-head gradient norm in WorldRoPE ($0.9\%$ at $692$ rotations/patch, $32$ scenes), where sub-patch gradients concentrate near $1.3$--$2.1$ rotations/patch ($20\%$).

\paragraph{Numerical precision at wide bands.}
To separate half-precision coordinate rounding from the analytic gradient mechanism of Eq.~(\mref{eq:box-grad}), we ran checkpoint probes and full 80K controls with single-precision camera geometry (\cref{tab:s-official-mods}, Fig.~\mref{fig:mechanism}(c)).
In half precision, a projected coordinate in the outer half of the image steps by $2^{-12}$ of the image width, corresponding to roughly five rotations at $3^{11}$ and $0.6$ at $3^9$.
This affects only the official codebase: our unified codebase already uses single-precision RayRoPE geometry (\cref{sec:s-code}) yet still loses $0.5$--$1.0\dB$ on RealEstate10K (\cref{tab:s-denc,tab:s-trained}) and $0.53\dB$ on ScanNet++ (\cref{sec:s-scannet}).

Checkpoint probes show that high-frequency gradients persist without rounding: switching a trained model to single-precision geometry at test time changes PSNR by only $-0.01\dB$, and frequencies $6$--$11$ carry almost the same depth-gradient share in full single precision ($51\%$ vs.\ $54\%$ as trained), even though half precision perturbs phase at the fastest frequencies.

Training from scratch with single-precision geometry eliminates mid-training loss jumps and narrows run-to-run spread from $1.85\dB$ to $0.28\dB$ across five runs ($0.23\dB$ across three seeds on B and C; \cref{tab:s-official-mods}).
This closes $60$--$63\%$ of the gap against the two jumped mixed-precision runs and $21$--$35\%$ against the three unjumped runs, yet every single-precision run still trails $243\times$ by $0.8$--$1.1\dB$.
Training the entire network in single precision gives nearly the same score, ruling out half precision elsewhere in the backbone, and wide-band degradation remains monotonic across $243\times$, $3^7$, $3^9$ ($-0.24\dB$), and $3^{11}$.

\begin{figure*}[t]
  \centering
  \includegraphics[width=0.86\textwidth]{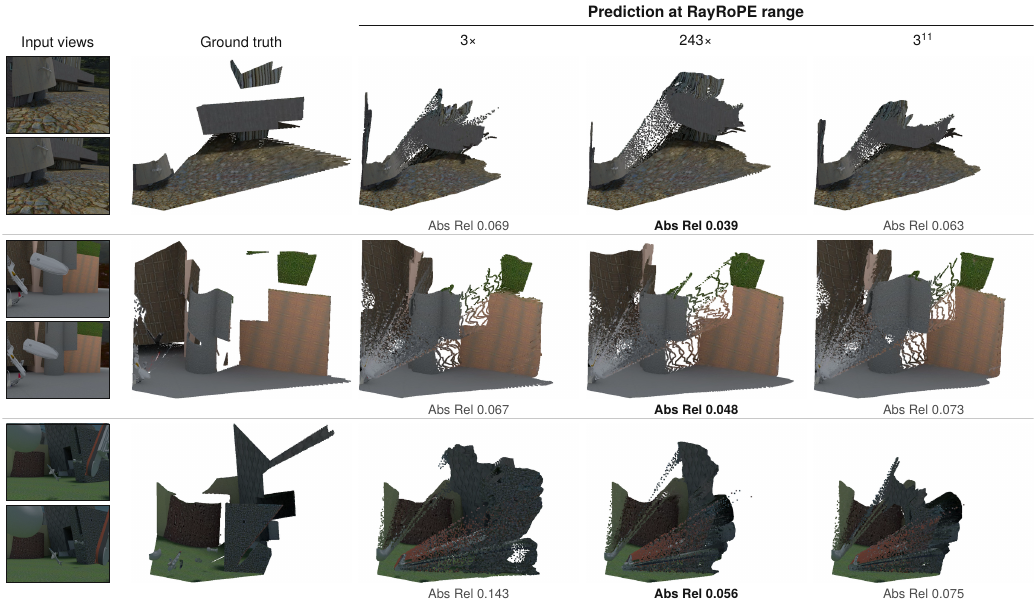}
  \caption{\textbf{How the frequency band changes 3D reconstruction on Scenes11.}
  Each row shows the two input images and four RGB point clouds rendered from the same oblique viewpoint in one common frame. We back-project the pixels inside each view's ground-truth evaluation mask using its calibrated camera and either ground-truth depth or the 50K-step UniMatch prediction at the indicated RayRoPE range (training seed 326), and combine the two clouds using the known relative pose. White denotes pixels with no projected samples. Numbers give Abs Rel averaged over the two ordered views (lower is better; best in bold). The three test sequences sit at the 73rd, 54th, and 98th percentiles of $243\times$-over-$3\times$ Abs Rel gains and the 87th, 90th, and 78th percentiles of $243\times$-over-$3^{11}$ gains across all 128 test sequences; over all of them, $243\times$ has the lower error on $84\%$ against $3\times$ and on $77\%$ against $3^{11}$.}
  \label{fig:s-scenes11-depth}
\end{figure*}

\paragraph{Combining single-precision geometry with stop-gradient.}
Adding high-frequency stop-gradient atop single-precision geometry recovers most of that remaining deficit across all four runs (\cref{tab:s-official-mods}): PSNR reaches $26.37$--$26.54\dB$ ($0.16\dB$ spread), adding $+0.70$ to $+0.87\dB$ over single-precision geometry alone and closing $76$--$92\%$ of the remaining gap ($85$--$94\%$ of the full gap; winning on over $90\%$ of scenes).
The two fixes are not additive: once high-frequency depth gradients are blocked, switching precision shifts PSNR by under $0.1\dB$ across matched pairs.
Using a machine-C $243\times$ reference changes seed-1 recovery shares by only a few percentage points.

\begin{table*}[t]
  \centering
  \caption{\textbf{Two-view depth estimation on Scenes11 across seeds} (per-seed results).
  Summary columns average the shared seeds; $\Delta$Abs Rel is paired per seed against the first row of each sub-block (in the $d=48$ block, single-ray $243\times$ is paired against the three-ray $3\times$ default; single-seed four-head rows are paired against seed~1 of their one-head baseline; values after semicolons report a third seed for the four one-head RayRoPE rows, giving three-seed means of $0.080$ vs.\ $0.096$ for three rays and $0.082$ vs.\ $0.089$ for one ray).}
  \label{tab:s-depth}
  \footnotesize
  \renewcommand{\arraystretch}{0.94}
  \begin{tabular}{@{}lrrcrrrll@{}}
\toprule
Encoding & Freq. & Range & Seeds & Abs Rel & RMSE & $\delta_1$ & Abs Rel per seed & $\Delta$Abs Rel per seed \\
\midrule
\multicolumn{9}{@{}l}{\emph{Single cross-attention head ($d{=}144$; main paper)}} \\
No encoding & -- & -- & 2 & 0.140 & 1.044 & 0.827 & 0.139 / 0.142 &  \\
\quad with 4 attention heads & -- & -- & 1 & 0.117 & 0.909 & 0.871 &  & $-$0.022 \\
\addlinespace
RayRoPE, 3 rays & 6 & $243\times$ & 2 & 0.077 & 0.635 & 0.941 & 0.086 / 0.068; 0.087 &  \\
RayRoPE, 3 rays & 6 & $3\times$ & 2 & 0.095 & 0.782 & 0.907 & 0.093 / 0.098; 0.096 & +0.007 / +0.030; +0.008 \\
\addlinespace
RayRoPE, 1 ray & 12 & $3^{11}$ & 2 & 0.084 & 0.675 & 0.932 & 0.085 / 0.084; 0.098 &  \\
RayRoPE, 1 ray & 12 & $243\times$ & 2 & 0.076 & 0.623 & 0.943 & 0.076 / 0.076; 0.093 & $-$0.009 / $-$0.008; $-$0.005 \\
\addlinespace
URoPE & 9 & $60\times$ & 2 & 0.070 & 0.578 & 0.953 & 0.068 / 0.072 &  \\
\quad with 4 attention heads & 9 & $60\times$ & 1 & 0.067 & 0.564 & 0.956 &  & $-$0.001 \\
\addlinespace
WorldRoPE & 24 & $68\times$ & 2 & 0.067 & 0.559 & 0.957 & 0.066 / 0.068 &  \\
\midrule
\multicolumn{9}{@{}l}{\emph{Three cross-attention heads ($d{=}48$, matching WorldRoPE's UniMatch setup; \cref{tab:s-published})}} \\
No encoding & -- & -- & 2 & 0.125 & 0.958 & 0.852 & 0.113 / 0.137 &  \\
RayRoPE, 3 rays (default) & 2 & $3\times$ & 2 & 0.097 & 0.783 & 0.905 & 0.100 / 0.094 &  \\
RayRoPE, 1 ray & 4 & $243\times$ & 2 & 0.079 & 0.648 & 0.940 & 0.078 / 0.081 & $-$0.023 / $-$0.013 \\
URoPE & 12 & $68\times$ & 2 & 0.067 & 0.563 & 0.956 & 0.065 / 0.069 &  \\
WorldRoPE & 8 & $68\times$ & 2 & 0.065 & 0.565 & 0.958 & 0.066 / 0.064 &  \\
\bottomrule
\end{tabular}

\end{table*}

\begin{table*}[t]
  \centering
  \caption{\textbf{Single-ray RayRoPE ($n=12$) frequency-range sweep on Scenes11} (numerical values for Fig.~\mref{fig:sweep}(d); two-seed means and per-seed $\Delta$Abs Rel vs.\ $243\times$, with third-seed values after semicolons).}
  \label{tab:s-depthrange}
  \footnotesize
  \renewcommand{\arraystretch}{0.94}
  \setlength{\tabcolsep}{5pt}
  \begin{tabular}{@{}lcrrrll@{}}
\toprule
Range & Seeds & Abs Rel & RMSE & $\delta_1$ & Abs Rel per seed & $\Delta$Abs Rel per seed \\
\midrule
$3^{1}$ & 2 & 0.098 & 0.790 & 0.901 & 0.096 / 0.101 & +0.020 / +0.025 \\
$3^{3}$ & 2 & 0.081 & 0.665 & 0.935 & 0.073 / 0.088 & $-$0.003 / +0.012 \\
$3^{5} = 243\times$ & 2 & \textbf{0.076} & \textbf{0.623} & \textbf{0.943} & 0.076 / 0.076; 0.093 & -- \\
$3^{7}$ & 2 & 0.078 & 0.636 & 0.941 & 0.077 / 0.078 & +0.001 / +0.002 \\
$3^{9}$ & 2 & 0.082 & 0.655 & 0.936 & 0.081 / 0.082 & +0.005 / +0.006 \\
$3^{11}$ & 2 & 0.084 & 0.675 & 0.932 & 0.085 / 0.084; 0.098 & +0.009 / +0.008; +0.005 \\
\bottomrule
\end{tabular}

\end{table*}

\paragraph{Cross-view attention concentration.}
For Fig.~\mref{fig:bandchart}(c), we average target-to-context attention across heads and the last three layers of official 25M models on 20 test scenes (three targets each), evaluated on \num{1927} SIFT~\cite{lowe2004sift} correspondences ($<1$-px reprojection error, $\ge 1^\circ$ parallax).
For each target token, we compute the fraction of its normalized attention in the matched context view falling within $r$ patches of the true match ($r=1$ covers the matched patch and its eight neighbors).
In every model, the attention peak lands within one patch on nearly all matches.
Yet $3\times$ RayRoPE places only $53\%$ of its weight within $r=1$ ($16\%$ on the exact match), versus $72\%$ ($26\%$ exact) for wider bands.
Compressing WorldRoPE or URoPE to $3\times$ causes a similar drop (from ${\sim}70$--$80\%$ down to ${\sim}57\%$), though wider-than-default bands also slightly reduce concentration and seeds vary by up to 11 percentage points, so we omit single-seed 47M comparisons.

\paragraph{Quality of internal predicted depth.}
Comparing last-layer predicted depths of official 47M models (20 scenes) against SIFT-triangulated depths via per-scene Spearman rank correlation (used because predicted depths have uncalibrated scale) shows consistent gains from stop-gradient across two seeds without relying on the jumped machine-A $3^{11}$ run ($0.36$).
On seed~1 (non-jumped $3^{11}$ on machine~B against $243\times$ and stop-gradient on machine~A), mean correlation is $0.46$ under $3^{11}$, $0.44$ under $243\times$ (paired $-0.02$ $[-0.05, 0.01]$), and $0.53$ under stop-gradient ($+0.07$ $[0.05, 0.09]$).
On seed~2 (all three models on machine~B), correlation rises from $0.31$ under $3^{11}$ to $0.44$ under $243\times$ ($+0.14$ $[0.10, 0.17]$) and $0.49$ under stop-gradient ($+0.19$ $[0.15, 0.22]$).
Because correlations do not match the PSNR order between $243\times$ and stop-gradient, and $243\times$ over $3^{11}$ holds on the same-machine seed~2 pair but not cross-machine on seed~1, this offers directional support alongside the main PSNR results.
In the analysis model (64 scenes), median last-layer log-depth error on context tokens likewise drops from $0.068$ ($3^{11}$) to $0.050$ ($243\times$).
Target tokens, whose depth is predicted without an input image, show larger errors of $0.4$--$0.6$ across layers ($0.39$ and $0.49$ in the last layer) with no consistent order.

\section{Depth Estimation}
\label{sec:s-depth}

\paragraph{Setup.}
Following prior UniMatch~\cite{xu2023unimatch} evaluations of RayRoPE~\cite{wu2026rayrope,podrope2026,worldrope2026}, we train single-scale UniMatch (six width-144 blocks, geometric RoPE in cross-attention only, 64-plane inverse-depth sweep) from scratch in single precision on DeMoN Scenes11~\cite{ummenhofer2017demon} for 50K steps (batch size 16, $224\times288$ crops of $240\times320$ images).
We evaluate all 128 test pairs in both directions (256 ordered pairs) at $240\times320$, upsampling predictions to $480\times640$ and scoring depths in $[0.5, 10]$.
Because this controlled recipe trains from scratch at lower resolution and fewer steps than full leaderboard schedules, absolute errors are higher; within each seed, all models share initialization and batch order.

\paragraph{Method comparisons across seeds.}
\Cref{tab:s-depth} gives the per-seed results behind Sec.~\mref{sec:other}.
Because inter-seed variance exceeds band gaps, paired within-seed comparisons are essential.
Every geometric RoPE outperforms the no-encoding baseline by a wide margin.
Within all three seeds, both narrowing three-ray RayRoPE ($243\times \to 3\times$) and widening single-ray RayRoPE ($243\times \to 3^{11}$) worsen Abs Rel across training checkpoints.
URoPE and WorldRoPE achieve slightly lower two-seed mean error than $243\times$ RayRoPE, though three-ray $243\times$ RayRoPE matches or beats them on seed~2.
Because standard UniMatch uses one head, we partition channels into four blocks for URoPE's four depth anchors; splitting cross-attention into four heads leaves URoPE unchanged while improving the no-encoding baseline.
To match WorldRoPE's published three-head UniMatch setting ($d=48$; \cref{tab:s-published}), the bottom block of \cref{tab:s-depth} splits cross-attention into three 48-channel heads (one depth anchor per head for URoPE).
Under this head size, RayRoPE's unchanged three-ray rule shrinks to two frequencies over $3\times$ and degrades sharply (two-seed mean Abs Rel $0.097$, vs.\ $0.065$ for WorldRoPE and $0.067$ for URoPE).
Switching to one ray with four frequencies over $243\times$ reduces Abs Rel by $-0.023$ ($[-0.029, -0.018]$) on seed~1 and $-0.013$ ($[-0.018, -0.008]$) on seed~2 (closing roughly half to two-thirds of the gap to WorldRoPE), reproducing the small-head narrow-band trap in two-view depth estimation.

\begin{figure*}[p]
  \centering
  \includegraphics[width=\textwidth]{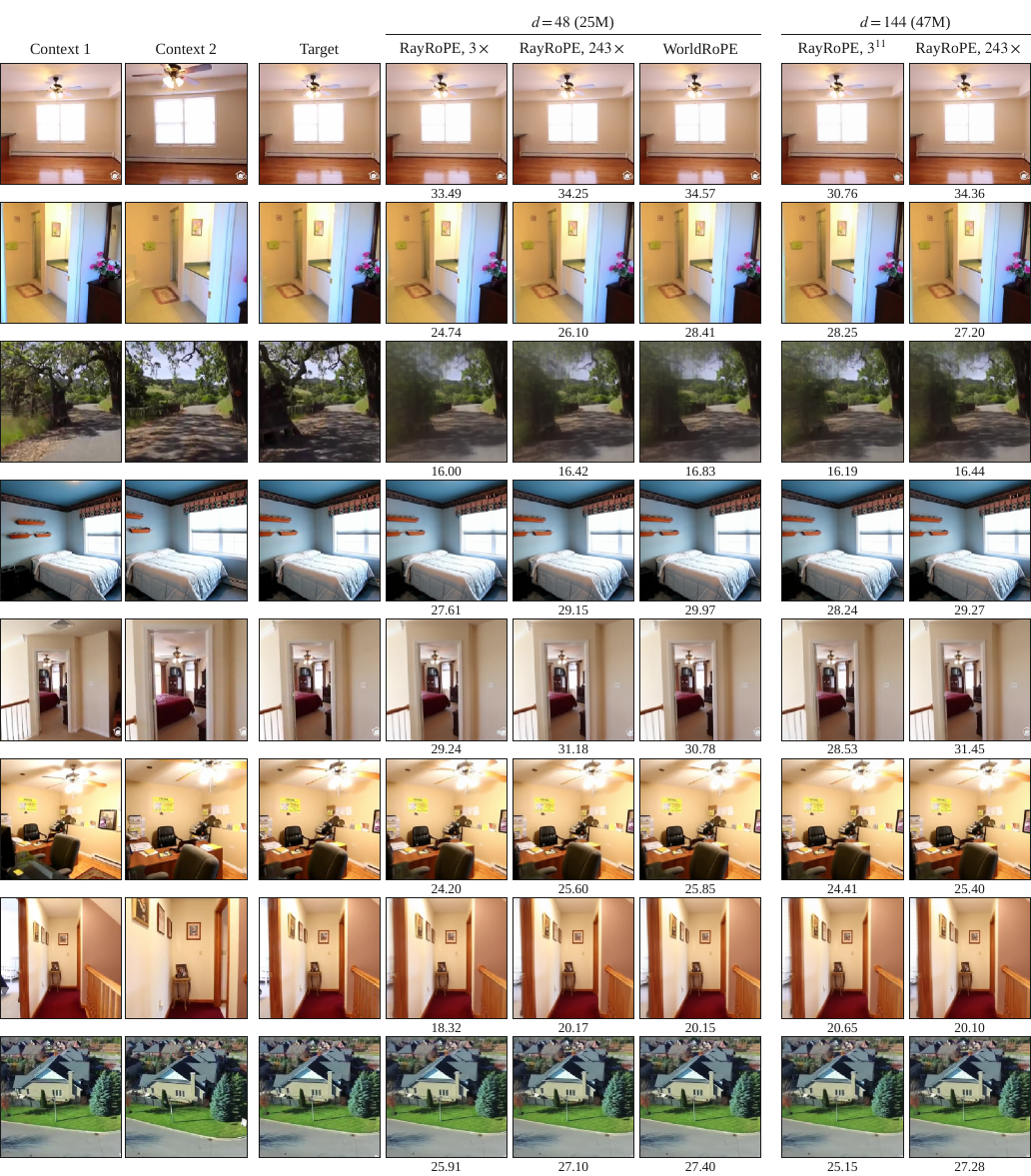}
  \caption{\textbf{Uncurated RealEstate10K test scenes} drawn at random (full middle target view and PSNR for each model in Fig.~\mref{fig:qual} except stop-gradient; the 25M $3\times$ column shows single-ray RayRoPE, whereas the webpage explorer shows the default three-ray model).}
  \label{fig:s-random}
\end{figure*}

\paragraph{Frequency-range sweep.}
Following Wu et al.~\cite{wu2026rayrope}, \cref{fig:s-scenes11-depth} back-projects both predictions into a common 3D frame, exposing boundary errors where misaligned transitions scatter points between surfaces.
\Cref{tab:s-depthrange} sweeps single-ray RayRoPE across the same six ranges as view synthesis; Fig.~\mref{fig:sweep}(d) plots the two-seed means and individual seeds against horizontal rotations per 8-pixel cell (on $224\times288$ training crops, where $\omega_0 = 1/288$ horizontally; the third seed, run only at $243\times$ and $3^{11}$, is omitted from the curve).

Mean Abs Rel is lowest at $243\times$, with $3\times$, $3^9$, and $3^{11}$ worsening error in every seed.
Scene-level 95\% bootstrap intervals (resampling the 128 scenes to account for paired view directions) exclude zero for every run except the third $3^{11}$ seed, which is borderline at the scene level and excludes zero across the 256 ordered pairs.
Nearby ranges remain close ($3^7$ stays within $0.002$ of $243\times$, while $27\times$ flips sign across seeds).
As in view synthesis, error rises much more steeply toward narrow bands ($+0.022$ at $3\times$) than toward wide ones ($+0.008$ at $3^{11}$).

\section{Additional Qualitative Results}
\label{sec:s-qual}

To illustrate failure modes clearly, the crops in Figs.~\mref{fig:teaser} and~\mref{fig:qual} were curated by hand.
For Fig.~\mref{fig:teaser}, we filtered the test targets for cases where $243\times$ peaks among the six ranges ($\ge 24\dB$), PSNR declines monotonically toward both extremes, and stop-gradient recovers $>2\dB$, picking one scene visually and cropping the $64\times64$ region of highest error contrast (top $1\%$ of $243\times$ gains at both ends).
In Fig.~\mref{fig:concept}, panels~(a) and~(b) are diagrams, while panel~(c) uses test scene \texttt{74e192d23f2fa8c4} (context view 0, target view 1) with a verified SIFT match: a $\pm15\%$ depth interval projects to a $0.45$-patch segment containing $0.005$, $0.43$, and $312$ cycles for $3\times$, $243\times$, and $3^{11}$ (drawn $77\times$ coarser for visibility).
In Fig.~\mref{fig:qual}, both targets sit around the 83rd--94th percentiles of $243\times$ gains over $3\times$ and $3^{11}$.
For an uncurated view of typical behavior, \cref{fig:s-random} shows the full middle target view of eight randomly sampled test scenes.
All $3^{11}$ renderings in Figs.~\mref{fig:teaser}, \mref{fig:qual}, and~\ref{fig:s-random} use the non-jumped machine-B checkpoint (\cref{tab:s-official-seeds}); for 25M RayRoPE at $3\times$, Figs.~\mref{fig:qual} and~\ref{fig:s-random} show the single-ray model, whereas the accompanying webpage shows the default three-ray model (so per-scene PSNRs for that column differ).

\end{document}